\documentclass[12 pt]{article}
\usepackage{times}
\usepackage{subfigure}

\usepackage{pifont}
\usepackage{floatrow}
\usepackage{caption}

\usepackage{mathrsfs}

\usepackage[fleqn]{amsmath}
\usepackage{amsfonts,amsthm,amssymb,mathrsfs,bbding}
\usepackage{txfonts}
\usepackage{graphics,multicol}
\usepackage{graphicx}
\usepackage{color}
\usepackage{caption}
\usepackage{indentfirst}
\usepackage{cite}
\usepackage{latexsym,bm}
\usepackage{enumerate}
\usepackage{epsfig}
\usepackage[noend]{algpseudocode}
 \usepackage[ruled,vlined]{algorithm2e}
\usepackage{tikz}
\usepackage{hyperref,todonotes,ulem,comment}
\usepackage{empheq,cases}

\usepackage{soul}

\evensidemargin=\oddsidemargin\topmargin=-1.5cm

\theoremstyle{definition}

\newtheorem{remark}{Remark}

\addtocounter{section}{0}

\def\H{{\mathcal{H}}}

\def\x{\bm{x}}
\def\X{\bm{X}}
\def\w{\bm{w}}
\def\y{\bm{y}}
\def\z{\bm{z}}
\def\v{\bm{v}}
\def\b{\bm{b}}

\def\@#1{{\cal #1}}
\def\d{\bm{d}}
\def\tu{\tilde{u}}
\def\dim{d}

\newcommand{\e}{ \mathbb{E}}
 \newcommand{\Cov}{\operatorname{Cov}}
 \newcommand{\var}{\operatorname{Var}}

\newcommand{\norm}[1]{\left\Vert#1\right\Vert}

\newcommand{\set}[1]{\left\{#1\right\}}
\newcommand{\Real}{\mathbb {R}}

\newcommand{\inpd}[2]{\left\langle #1, #2 \right\rangle}
\newcommand{\tr}{\textsf{T}}

\begin{document}
\title{Active Learning for Saddle Point Calculation}

\author{Shuting Gu$^a$,\quad Hongqiao Wang$^{b*}$\quad
{and} \quad
Xiang Zhou$^c$ \\
{\it $^a$ College of Big Data and Internet} \\
{\it Shenzhen Technology University} \\
{\it Shenzhen 518118, People’s Republic of China} \\
[2mm]
{\it $^b$School of Mathematics and Statistics} \\
{\it Central South University} \\
{\it Changsha 410083, People’s Republic of China}\\
[2mm]
{\it $^c$School of Data Science and Department of Mathematics}\\
{\it  City University of Hong Kong}\\
{\it Tat Chee Ave., Kowloon, Hong Kong SAR, China}\\
[2mm]
{$^*$ Corresponding author: Hongqiao Wang}\\
{\it School of Mathematics and Statistics} \\
{\it Central South University} \\
{\it Changsha 410083, People’s Republic of China}\\
{\it E-mail: Hongqiao.Wang@csu.edu.cn}}

\date{}

\maketitle {\flushleft\large\bf Abstract }
The saddle point (SP) calculation is a grand challenge for computationally intensive energy function in computational chemistry area, where   the saddle point may represent the transition state (TS). The traditional methods need to evaluate the  gradients of the energy function at a very large number of locations. To reduce the number of expensive computations of  the true gradients, we propose an active learning framework consisting of   a statistical surrogate model, Gaussian process regression (GPR) for the energy function,  and a single-walker dynamics  method, gentle accent dynamics (GAD), for the saddle-type transition states. SP is detected by the GAD applied to the GPR surrogate for the gradient vector and the Hessian matrix. Our  key ingredient for efficiency improvements is an active learning method which sequentially designs  the most informative locations and takes evaluations of  the original model at these locations to train GPR.  We formulate  this active learning task as the optimal experimental design problem  and propose a very efficient  sample-based   sub-optimal criterion to construct the optimal locations. We show that the new method significantly decreases the required number of  energy or force evaluations of the original model.




\section{Introduction}

The investigation of reaction mechanisms is a central goal in theoretical chemistry. 
Any reaction can be characterized by its potential energy surface (PES), the energy depending on the nuclear coordinates of all atoms.
Minima on the PES correspond to reactants and products. 
The minimum-energy path (MEP), the path of lowest energy that connects two minima, can be seen as an approximation to the mean reaction path.
It proceeds through a transition state (TS),  a special type of the saddle point (SP) with index-1, which is defined as the state with exactly one dimensional unstable manifold, or the unstable critical point whose Hessian has only one negative eigenvalue. 
The energy difference between a minimum and a TS connected to the minimum is the reaction barrier, which can be used in transition state theory\cite{pechukas1981transition,truhlar1996current} to calculate the reaction rate constants.

Therefore,  locating the SP on PES is an important topic in computational chemistry and other related applied fields. 
 A large number of numerical methods have been proposed and developed to efficiently compute the SP. 
Generally speaking, there are two classes: path-finding methods and surface-walking methods.
The former includes the string method \cite{String2002}, the climbing string method\cite{Ren2013} and the nudged elastic band method\cite{jonsson1998nudged}.
These methods are to search the so-called MEP. The points along the MEP with locally maximum energy value are then the SP.
The latter methods include the eigenvector-following method\cite{Crippen1971}, the dimer method\cite{Dimer1999}, the activation-relaxation techniques\cite{ART1998}, the gentlest ascent dynamics (GAD)\cite{GAD2011,SamantaGAD},  the iterative minimization formulation\cite{IMF2014,IMA2015} and others \cite{zhang2016optimization,yu2021global}.
They evolve a single state on the potential energy surface by using the unstable direction, for example, the min-mode direction.
All algorithms in  the two classes rely on the iterative evaluation of the  force vector (gradient) of PES (as well as  Hessian matrix in some methods).
Therefore, the required number of force evaluations is an important performance indicator in comparing various numerical methods for saddle points.

Force evaluations can be  prohibitively  expensive for   large-scale models.
One typical example is the free energy surface where each evaluation on the free energy surface requires a long time molecular dynamics simulation. 
Thus  the aforementioned traditional methods are  time-costing if applied directly.
 In order to boost the efficiency, recently there have been significant works to build machine learning (ML) surrogate models
 \cite{chen2018atomic, BASC2016, khorshidi2016amp, NEB_GPR, NEB_GPR_PRL, guan2018construction, lin2021searching, RiD2018}.
 The idea of  these methods is to construct a good surrogate function which closely approximates the target potential function in the region of interest.  
 
 However, there are several different scenarios  of using the surrogate models, which will determine how to efficiently select the data samples in training surrogate models,
 i.e., the  query method in active learning.   
  One of main scenarios is to explore the whole configuration space by overcoming rare events and various barriers \cite{chen2018atomic, khorshidi2016amp, lin2021searching, guan2018construction,RiD2018},
 then one should train a surrogate model to represent the overall quality of the free energy almost everywhere, 
 by carefully handling   meta-stable states. 
 In general,  the ``uncertainty''-based rule of active query \cite{settles2010active} is usually adopted in these works \cite{RiD2018,lin2021searching}, 
 and   the specific definition of  ``uncertainty'' concept can vary case by case based on  various heuristics. 
 If the purpose of constructing surrogate models is only to look for the global minima on potential energy surfaces\cite{BASC2016}, 
 then the idea of  the Bayesian optimization (BO) is applicable by
 maximizing {\it expected improvement} as one popular approach of query to generate  new data point.   However, BO can not handle the SP search problem directly  because SP problem is actually a max-min problem in mathematics.
For the purpose  of searching the SP with the aid of surrogate models, the recent works in 
\cite{NEB_GPR, NEB_GPR_PRL,denzel2018gaussian} combine  the   traditional  path-finding  methods
 (e.g., nudged elastic band method \cite{jonsson1998nudged}) with the Gaussian process regression (GPR),   where   the path between  two given local minima is the main object 
 and the prediction of the GPR  with fixed hyper-parameters is updated sequentially with new data.
 The   new data positions   are selected mostly based on some intuitions such as the local energy-maximal points,
 which proved quite efficient  in experiments. 
Yet,  the path-based methods rely  on the prior information of two different local minima and a good initial path.
 
In this paper, we propose a rational  strategy of active learning and a practical algorithm based on Gaussian process regression  and a single-walker dynamics to reduce  the number of  expensive evaluations in  SP calculation. 
The SP could be calculated by any dynamic model of searching saddle points and here we choose the gentle accent dynamics (GAD)\cite{GAD2011} with consideration of the applicability to both gradient and non-gradient systems.  The GAD uses the gradient vector and estimates the Hessian matrix of the PES in a local domain.
 In each GAD iteration, we use the derivatives estimated by the Gaussian process regression  instead of the real ones of the PES.
We design a meaningful and simple  mechanism to determine whether the current GPR is reliable or not to carry on the current GAD iteration. Then 
we propose an active learning framework to select a new batch of data samples to sequentially update the parameters in the surrogate model. 
The locations of the new data are determined by optimal experimental design theory.
We call this method as the {\it adaptive GPR-GAD} (aGPR-GAD).
We propose a systematic and rational design criterion starting  from the traditional  expected utility function, which is based on the mutual information between the data locations and the predicted GAD trajectories. 
This criterion is implemented in a sample-based approach so that it  is computationally efficient and can avoid the notorious density estimation in the classical optimal design criterion. 
The results of our three numerical examples show that this new method not only accurately and robustly locates the SP but also efficiently reduces the number of evaluations compared with the classical GAD. 


The rest of the paper is organized as follows. We first review the preliminaries of our work in Section \ref{review}, including the gentlest ascent dynamics and the Gaussian process surrogate model for the derivative estimation. In Section \ref{main_result}, we introduce the main idea of our aGPR-GAD method for searching saddle points.
A general experimental design framework and a novel optimal criterion derived from mutual information are presented in Section \ref{sec:active_learning}.
Numerical examples are presented in Section \ref{ex} to demonstrate the effectiveness of the proposed method, and Section \ref{conclusion} offers some concluding remarks.

\section{Background}\label{review}

\subsection{Gentlest Ascent Dynamics}
\label{set:GAD}
 Gentlest ascent dynamics (GAD)  \cite{GAD2011,MsGAD2017,simGAD} is a  continuous dynamics  for searching saddle points.
For  the  index-1 saddle point of the dynamical system  $\dot{\x}(t) = \b(\x)$ in $\mathbb{R}^d$, 
the classical GAD  couples the position variable $\x$ and the two direction variables $\v$ and $\w$ as follows:
\begin{subequations}\label{GAD}
\begin{empheq}[left=\empheqlbrace]{align}
  \dot{\x}(t) &= \b(\x) -2 \frac{\inpd{\b(\x)}{ \w }}{\inpd{\w}{\v} }\v,\label{GAD-x}\\
     \dot{\v}(t) &=  J(\x)\v - \alpha \v, \label{GAD-v}\\
   \dot{\w}(t)&=  J(\x)^\tr \w - \beta \w,  \label{GAD-w}
\end{empheq}
\end{subequations}
where $J(\x) = D \b(\x)$ is the Jacobian matrix $\left(D \b\right)_{ij} :=
\frac{\partial b_i}{\partial x_j}$.
$\alpha$ and $\beta$ are the Lagrangian multipliers to impose
certain  normalization conditions for $v$ and $w$. For instance,
if the normalization condition is  $\inpd{\v}{\v}\equiv \inpd{\w}{\v}\equiv 1$, then   $\alpha=\inpd{\v} {J(\x)\v}$ and $\beta=2\inpd{\w}{J(\x) \v}-\alpha$.  Equation \eqref{GAD}
is a flow in $\Real^{3d}$.
In \cite{simGAD}, the  GAD \eqref{GAD}  is simplified  and has  only one direction variable;
for instance, if we use $\v$, then 
\begin{subequations}{\label{HGADv}}
\begin{empheq}[left=\empheqlbrace]{align}
 \dot{\x}  &=  \b(\x)-2 \inpd{\b(\x)}{\v(t)} \v(t) / \norm{\v(t)}^2,
 \\
   \dot{\v}  &=  J (\x) \v - \inpd{\v}{J (\x) \v} \v.
   \end{empheq}
\end{subequations}

As a special case, the GAD for a gradient system $\dot{\x}(t) = -\nabla u(\x)$, where $u\in C^2(\Real^d)$ is the potential energy function, is
\begin{subequations}{\label{GAD-g}}
\begin{empheq}[left=\empheqlbrace]{align}
  \dot{\x}(t) &= -\nabla u(\x) + 2 \frac{\inpd{\nabla u(\x)}{ \v}}{\inpd{\v}{\v}} \v,\label{GAD-g-x}\\
   \dot{\v}(t)  &=   - \nabla^2 u(\x)\v + \inpd{\v} {\nabla^2 u(\x)\v}\v. \label{GAD-g-v}
   \end{empheq}
\end{subequations}
 For a frozen $\x$, the steady state of the direction $\v$ is the min mode of the Hessian matrix $\nabla^2 u(\x)$:
 the eigenvector corresponding to the smallest eigenvalue of $\nabla^2 u(\x)$. Equation \eqref{GAD-g} converges to the saddle point of the potential $u(\x)$.

 In our paper,  we mainly work with the GAD for the gradient system, i.e., equation \eqref{GAD-g},
 but we also discuss  how to generalize to the non-gradient system. 
 So we adopt  the dynamic form \eqref{HGADv} and work with $\b$ and $J$.
  In gradient systems, we start from the energy function $u$, then compute 
  the derivative and the Hessian matrix 
  $\b(\x)=-\nabla u(\x)$,   $J(\x)=-\nabla^2 u(\x)$.
  In non-gradient systems, we directly start with the vector field $\b$.
    
 The  time-discrete form of \eqref{HGADv} in the Euler scheme is  as follows with the time step size $\Delta t$ and $t=1,2,3,\dots$,
\begin{equation}{\label{HGADv_dis}}
\begin{split}
 \x^{(t)}  &= \x^{(t-1)} + \Delta t ~
 \left( \b(\x^{(t-1)})-2 \inpd{\b(\x^{(t-1)})}{\v^{(t-1)}} \v^{(t-1)} / \norm{\v^{(t-1)}}^2\right),
 \\
   \v^{(t)}  &=\v^{(t-1)} + \Delta t ~\left(  J (\x^{(t-1)}) \v^{(t-1)} - \inpd{\v^{(t-1)}}{J(\x^{(t-1)})  \v^{(t-1)}} \v^{(t-1)} \right).
\end{split}
\end{equation}
 
\subsection{Review of Gaussian process and derivative estimation}
\label{set:GPR}
The Gaussian process regression or Kringing method constructs the approximation of a function   in  a non-parametric Bayesian regression framework \cite{rasmussen2003gaussian}. 
The Bayesian inference is based on both prior and observed data. A characteristic of the Bayesian method, including Gaussian process regression, is that this method tolerates a coarse assumption of the prior distribution. In addition, the impact of specific prior distribution becomes less important by maximizing the marginal likelihood function with more and more observed data. In our problem of free energy surface, free energy is a macroscopic model, coarse-grained from molecular dynamics simulation, so the functions to approximate are pretty smooth in practice and are well suitable for applying the Gaussian process regression.

We illustrate the technique of GPR by 
considering the energy function $u(\x)$ as the target function to approximate. 
Specifically  we use a  Gaussian random process  $\tilde{u}(\x)$ as a {\it surrogate model}  
for the   function $u$.
First of all, before   any data from $u$ is available, $\tilde{u}(\x)$ is assumed to have the prior distribution  with zero expectation: 
\begin{equation}
\label{eq:u_GP}
\tu(\x) \sim \mathcal{GP}\left(0, \Cov[\tu(\x),\tu(\x')]\right),
\end{equation}
where   the covariance function
\begin{equation}
\label{eq:COV}
 \Cov[\tu(\x),\tu(\x')] = k_{u,u}(\x,\x')
\end{equation}
 is specified by a positive semidefinite and bounded  kernel function
  $k_{u,u}(\x,\x')$. The subindex pair
here in the kernel refers to the underlying target function $u$ in consideration.
We assume that the kernel $k_{u,u}$ is sufficient smooth 
and satisfies certain growth condition \cite{adler1981geometry}
such that the sample path $\tilde{u}(\x)$ is  in $C^2$ with probability one.
We choose the following 
squared exponential kernel in our paper to have the $C^2$ property of the sample path:
\begin{equation}
\label{eq:app1}
  k_{u,u}(\x, \x') =  \eta \exp(-\frac{\|\x - \x'\|^2}{2 l}),
\end{equation}
where $||\cdot||$ is the $l_2$ norm, $\eta$ and $l$ are two positive hyper-parameters.

 Here we are interested in   the first order (vector-valued) derivative
  and the second order (matrix-valued) derivative of $\tu(\x)$, which are denoted by  $$\tilde{\b}\triangleq (\tilde{b}_i)=-\nabla \tu(\x)
  \quad \mbox{ and } \quad \tilde{J}\triangleq (\tilde{J}_{ij}) =-\nabla^2\tu(\x),$$ respectively. 
A key property of the Gaussian process in our favor is that any linear transformation, such as   differentiation and   integration, of a Gaussian process  is still a Gaussian process \cite{adler1981geometry,rasmussen2003gaussian, solak2003derivative,wang2021explicit}. 
For example,  for each $i,j$, the functions $\tilde{b}_i(\cdot)$ and $\tilde{J}_{i,j}(\cdot)$ are also mean-zero Gaussian processes: 
$\tilde{b}_i  \sim \mathcal{GP}( {0}, k_{b_i,b_i}(\x,\x')),   
\tilde{J}_{i,j}   \sim \mathcal{GP}( {0}, k_{J_{i,j},J_{i,j}}(\x,\x')),
$ with the covariance functions $k_{b_i,b_i}(\x,\x')=\frac{\partial^2 }{\partial x_i x'_i} k_{u,u}(\x,\x')$ and  $k_{J_{i,j},J_{i,j}}(\x,\x')=\frac{\partial^2 }{\partial x_ix_j } \frac{\partial^2 }{ \partial x'_ix'_j}k_{u,u}(\x,\x') $, respectively.
More generally, one can write the joint prior distribution for the $(1+d+d^2)$-dim  
 Gaussian random function $(\tu, \tilde{\b}, \tilde{J})$.

Next, we consider the observation data and the posterior distributions of $(\tu, \tilde{\b}, \tilde{J})$. 
In practice, the noisy energy function value are measured and   assumed to be the random samples of the model at certain locations, 
\begin{equation}\label{261}
y(\x) =\tu(\x) + \epsilon(\x),
\end{equation}
where the measurement noise $\epsilon(\x)$ is a zero-mean Gaussian random function   with 
the Dirac delta covariance function $\Cov(\epsilon(\x),\epsilon(\x'))=0$ if $\x \neq \x'$ and the variance  $ \var(\epsilon(\x)) \equiv \sigma^2_u$.
It is clear that $y$ is a Gaussian process $\mathcal{GP}(0, k_{u,u}(\cdot,\cdot) + \sigma_u^2 \delta(\cdot,\cdot))$.

The set of locations where the observation is measured is denoted by  $\X^* := \left[\x^*_1, \cdots \x^*_{n}\right]^\mathrm{T}$,
where  $n$ observation data $Y^* := \left[ {y}^*_1, \cdots, {y}^*_{n}\right]^\mathrm{T}$  of the energy function $u$  are measured,  assumed to follow the rule \eqref{261}.
$\X^*$  is often known as the training location points 
and $Y^*$ is their corresponding labels. 
 
Conditioned  on the given data-set denoted by $\mathcal{D}=\{\X^* ,Y^*\}$, 
we are interested in the Gaussian process regression (GPR),
the task of predicting  $\b(\x)$ and $J(\x)$ at any   test location $\x\in \Real^d \setminus  \X^*$.
This is achieved by  
considering the   posterior distribution of 
the $\mathbb{R}^{(d+d^2)}$-valued Gaussian random function $$\mathcal{F}(\x)\triangleq [\tilde{\b}(\x), \tilde{J}(\x)]^\mathrm{T} = [\tilde{b}_1,\dots,\tilde{b}_d,\tilde{J}_{11},\dots,\tilde{J}_{ij},\dots,\tilde{J}_{dd}]^\mathrm{T},$$
which has the following distribution  \cite{rasmussen2003gaussian, solak2003derivative}:
\begin{equation}
\label{eq:post_b}
\mathcal{F}(\x)~|~\x, \mathcal{D}~\sim \mathcal{N}(\bm{\mu}^\mathcal{F}(\x), \Sigma^\mathcal{F}(\x) ),
\end{equation}
where 
\begin{align}
\label{eq:postMean}
 \bm{\mu}^\mathcal{F} &= [\mu^{b_1},\dots,\mu^{b_d},\mu^{J_{11}},\dots,\mu^{J_{ij}},\dots,\mu^{J_{dd}}]^{\mathrm{T}}= K_{\mathcal{F},u}(\x,\X^*)K_{u,u}(\X^*,\X^*)^{-1}Y^*,\\
 \label{eq:postCov}
\Sigma^\mathcal{F} &= K_{\mathcal{F},\mathcal{F}}(\x,\x)-K_{\mathcal{F},u}(\x,\X^*)K_{u,u}(\X^*,\X^*)^{-1}K_{u,\mathcal{F}}(\X^*,\x).
 \end{align}
with \begin{equation}
\label{eq:jointdis2}
    K_{u, \mathcal{F}} = [ K_{u,\b}(\X^*,\x), K_{u,J}(\X^*,\x)],\quad 
    K_{\mathcal{F}, \mathcal{F}} =\left[
				 \begin{array}{llll}
          K_{\b,\b}(\x,\x) & K_{\b,J}(\x,\x) \\
         K_{J,\b}(\x,\x) & K_{J,J}(\x,\x)
         \end{array}\right],
\end{equation}
and $K_{\mathcal{F}, u} = K_{u, \mathcal{F}}^\mathrm{T}$.   
The specific expressions of these matrices  can be written in terms of 
the kernel $k_{u,u}$  (Appendix \ref{set:appgpr}).
The size of the positive definite matrix $K_{u,u}(\X^*,\X^*)$ is $n$ by $n$, where $n$
is the number of  the sampled locations  in $\X^*$.
The hyper-parameters $\eta$ ,  $l$  and $\sigma_u^2$ are trained by the standard  maximum likelihood estimation (MLE)  of maximizing the marginal likelihood $p(Y^*|\X^*, \eta, l, \sigma_u^2)$ with the given data-set $\mathcal{D}$.
See the details in  \cite{rasmussen2003gaussian}.

\begin{remark}
For the non-gradient system \eqref{HGADv}, the above Gaussian process regression can also be used to  approximate  the force field $\b(\x)$ with the  direct observations   $\b(\x_i^*)$. In the example \ref{set:non_gradient}, we will encounter this situation.
It is also well noted that in most free energy calculation methods\cite{Maragliano2006StringMI,branduardi2007b}, the measured data is the so called {\it mean force}, which is the noisy approximation of $\nabla u$.
In both cases,  one then only need the estimation of ``first-order derivative" of the observed data, which is  the Jacobi matrix $D \b(\x)$.
Accordingly, the expression of \eqref{eq:post_b} should be modified. 
To be flexible,  our formulation through the paper is still  for the observation of the potential function $u$.
\end{remark}

\section{Surrogate model-based GAD}\label{main_result}
As aforementioned   in Section \ref{set:GAD}, GAD is a dynamic method for locating the SP  based on the local information: the negative gradient vector $\bm{b}$ and the negative Hessian matrix $J$ of the energy function.
In practice, the energy function and its corresponding derivatives do not admit analytical expression and have to be evaluated through computational intensive molecular dynamics simulations. 
If we had some observations of the energy function values $Y^* $ at the   locations $\X^*$, a surrogate models for $\tilde{\b}$ and $\tilde{J}$ can be constructed by the GPR  with the given dataset $\mathcal{D}  = \{\X^*, Y^*\}$, as introduced in Section \ref{set:GPR}.
We then can   approximate the first and the second derivatives of the energy function by their mean functions $\mu^{b_i}$ and $\mu^{J_{ij}}$ of the GPR surrogate model in \eqref{eq:postMean}. Then the gradient vector and the Hessian matrix in \eqref{HGADv} can be replaced by 
\begin{equation}
\label{eq:surrogate}
\begin{split}
\bm{\mu}^b(\x) &= [\mu^{b_1}(\x),\dots,\mu^{b_d}]^\mathrm{T}\quad
\text{and} \quad  \bm{\mu}^J(\x) &=\left[
				\begin{array}{lll}
         \mu^{J_{1,1}}(\x) &\cdots &\mu^{J_{1,d}}(\x) \\
         \cdots&\cdots& \cdots\\
        \mu^{J_{d,1}}(\x)  &\cdots&\mu^{J_{d,d}}(\x)
        \end{array}\right].
\end{split}
\end{equation} 
 If the surrogate model is accurate, the SP can be easily detected by the GAD with  \eqref{eq:surrogate} and without any more expensive MD simulation.
But as we know, in order to train an accurate machine learning model, sufficient data is vital. Though Gaussian process regression is an outstanding model in small dataset, it still needs lots of data for training an accurate mean function in \eqref{eq:surrogate} in the whole domain, which is still computational intensive. 
In  \eqref{HGADv}, SP can be detected by the GAD as  the terminal  point of the  GAD trajectory.
 Thus the SP detection problem can be replaced by the problem of detecting the trajectory of GAD.
Besides, the GAD is a local method which means only local derivative information is sufficient.
So we only need the surrogate function to be  accurate around  the GAD trajectory instead of the whole domain. 

We propose to update the surrogate model sequentially for maintaining sufficient prediction accuracy at the points where the GAD marches.
The surrogate function is trained by the Gaussian process regression and updated by adding new data into the training set. 
The new design locations  will be determined by the active learning strategy in Section \ref{sec:active_learning}. 
The observations at the new data locations are evaluated through computational intensive access to the true energy function. Specifically, we solve the dynamic system \eqref{HGADv}   by replacing the true gradient vector and Hessian matrix by the mean functions in \eqref{eq:surrogate}. With the limited observations, the dynamic solution would arrive at a location where the uncertainty of the surrogate model is too large and the estimators \eqref{eq:surrogate} become unreliable.
We assume this happens at  $\x^{(t)}$ and $\v^{(t)}$ and we next shall add a batch of  $N_{D}$ new data.
The locations of the new data are called {\it design locations} (also called {\it query points, experiment conditions}) and denoted by $D = [\d_1, \dots,\d_{N_D}]\in \Real^{d\times N_{D}}$,
 $\d_j \in \Real^d$. 
The design locations are essential to maximally reduce the uncertainty of the surrogate model. 
We hope to decide the design locations which can offer the most information for finding the future trajectory of the GAD system.  
An active learning strategy can serve this purpose and it will be introduced in detail in Section \ref{sec:active_learning}.
The new data labels are obtained by taking evaluations at the optimal design locations.
Hence a sequential design strategy for updating the surrogate model is naturally formed and incorporated into the GAD iteration.
The main algorithm  is described in Algorithm \ref{alg:gad-gpr1}. 
  \IncMargin{1em}
\begin{algorithm}[H]\label{alg:gad-gpr1}
\caption{aGPR-GAD method}
\LinesNumbered 
\SetKwInOut{Input}{Parameter}\SetKwInOut{Output}{Output}
\SetAlgoLined
\Input{max  iteration $t_{\max}$, starting point  and its corresponding direction $\{\x^{(0)}, \v^{(0)}\}$, time step $\Delta t$; number of initial data points $N_0$ and number of design locations  
in each active learning update $N_D$;} 
\Output{Transition state $\x^{SP}$; }
Randomly select $N_0$ points $(\x_1,\dots,\x_{N_0})$ around $\x^{(0)}$ and evaluate their corresponding energy function values $(y_1,\dots,y_{N_0})$. Obtain the initial data-set  $\mathcal{D} = \{ \x_i, y_i\}_{i=1,\cdots,N_0}$\;
Train the GPR surrogate  function $\mathcal{F}$ and get the hyper-parameters $\eta$ ,  $l$  and $\sigma_u^2$ by MLE with the data-set $\mathcal{D}$  (Section \ref{set:GPR})\;
\For{$t=1$ \KwTo$t_{\max}$} 
{Compute $\x^{(t)}$ and $\v^{(t)}$ by Equation \eqref{HGADv_dis} with time step $\Delta t$, the gradient vector and Hessian matrix computed by $\bm{\mu}^b(\x^{(t-1)})$ and $\bm{\mu}^J(\x^{(t-1)})$ in Equations \eqref{eq:surrogate} and \eqref{eq:postMean}\;
\If  {$||\x^{(t)} - \x^{(t-1)}|| + ||\v^{(t)} - \v^{(t-1)}|| <tolerance $} 
{$\x^{SP} = \x^{(t)}$ ; Break the loop;}
\If { $\mathcal{F}$ is unreliable at $\x^{(t)}$ (equation \eqref{eq:ths})} 
			{Find $N_D$ new design locations $D=\{\bm{d}_1, \dots ,\bm{d}_{N_{D}} \}$  by active learning strategy using Algorithm 2 in Section \ref{sec:active_learning}\;
			Evaluate the  corresponding   values $\{y_1, \dots ,y_{N_{D}} \}$ at $D$ by accessing the original model $u(\x)$\;
			Use the new data-set $\mathcal{D} = \mathcal{D} \cup \{(\bm{d}_i, y_i)\}_{i=1,\dots,N_D}$ to retrain  the GPR surrogate $\mathcal{F}$ and update the hyper-parameters $\eta$ ,  $l$  and $\sigma_u^2$ by MLE (Section \ref{set:GPR}) \;
			 }
}

\end{algorithm}
\DecMargin{1em}

The criterion of ``$\mathcal{F}$  is unreliable at $\x^{(t)}$" in Line 8 is  proposed below.   
Intuitively the variance values of $\mathcal{F}(\x^{(t)})$ computed by \eqref{eq:postCov}, is a reasonable indicator for quantifying the  uncertainty of the surrogate model.
Note here $\mathcal{F}$ includes both $\b$ and $J$; to   balance the variance of $\tilde{\bm{b}}(\x^{(t)})$ and $\tilde{J}(\x^{(t)})$,
we will learn a linear combination  of  $\b$ and $J$ (which will be discussed in Section \ref{sec:active_learning})
and  consider  the  quantity  $||\alpha \bar{\Sigma}^b + \bm{\beta}\bar{\Sigma}^J||_{L_\infty}$ where
\begin{equation*}
\bar{\Sigma}^b = [ \text{Var}(b_1), \dots, \text{Var}(b_d)]^{\mathrm{T}} \quad 
\text{and}
\quad
 \bar{\Sigma}^J =
 \begin{bmatrix}
  \text{Var}(J_{1,1})&\dots & \text{Var}(J_{1,d})\\
 \vdots&  &\vdots\\
  \text{Var}(J_{d,1})&\dots & \text{Var}(J_{d,d})
 \end{bmatrix}.
\end{equation*}
The algorithm sets the surrogate function $\mathcal{F}$ as  unreliable whenever
\begin{equation} \label{eq:ths}
||\alpha \bar{\Sigma}^b + \bm{\beta}\bar{\Sigma}^J||_{L_\infty} \geq  \sigma^2_{sur},
\end{equation}
 where $\sigma^2_{sur}$ is a prescribed     threshold.
After we introduce the main framework, we next develop the active learning strategy for design positions.

%
%

\section{Active learning for the design locations}
\label{sec:active_learning}
Active learning, which is also called ``optimal experimental design" in statistics, is a sub-field of machine learning and, more generally, artificial intelligence \cite{settles2010active}.
The key idea behind active learning is that a machine learning algorithm can achieve greater accuracy with fewer labeled training data if it is allowed to choose the sample from which it learns.
Therefore it is a very powerful tool for learning problems where sample labeling is very difficult, time-consuming or expensive and  it has achieved great success in some fields, such as the speech recognition \cite{ IEEE2005, Allen2002}, the information extraction\cite{Jones03activelearning} and the natural language processing \cite{Olsson2009}.
The required freedom of being able to select the query points arbitrarily    in active learning   is naturally  satisfied in computer simulations for scientific computing problems.  There have been some related studies\cite{lin2021searching,guan2018construction,uteva2018active,smith2018less,lin2020automatically,zhang2019active} in the field of computational chemistry.
Active learning shows a huge advantage in reducing the labeled data size and it is well suitable for our saddle point calculation problem.

\subsection{Active learning as Bayesian experiment design}
The assumption of active learning is that one can  access to a true black-box model which is expensive to evaluate 
and  also can train a data-driven machine learning  model (i.e., surrogate model)  based on the finite samples of data.
In contrast to many traditional machine learning problems with a fixed set of  training samples, 
the surrogate model in active learning  is recursively retrained  
by sequentially adding more and more samples into the training set.
The main task of active learning is that
based on an existing surrogate machine learning model, 
how to find a good set of  new design locations $D $, so that 
after  updating the current machine learning model  by 
using the new   observation data $\y^* = (y^*_1,\dots,y^*_{N_D})$  
at  $D$,  the retrained  model is more accurate in certain sense than a blind draw of the new location.
In this way, the value of each data sample is maximized and the number of accessing the time-consuming experiments
is much reduced.

First of all, we need introduce a   parameter of interest,  denoted by $Z$,   which  is the ultimate goal of the surrogate model  used to infer.
In general,   $Z$ refers to the unknown parameter of the inference in the surrogate model, which was denoted as  $\boldsymbol{\theta}$ in \cite{huan2013simulation}.
 Specifically for our own problem here, $Z$ refers to  the  trajectory $\x^{(t)}$ of the GAD from the initial point to a saddle point.
 If this trajectory $\x^{(t)}$ {\it was} known,  we only need to access the original model  on {\it this} trajectory, instead of the whole space.

To evaluate the performance of any strategy of experiment design, 
one can examine the difference between the ground truth of $Z$ and the inferred value of $Z$ of a machine learning model sequentially  trained  by this strategy.
To be tractable in computation, 
determining the optimal location set $D$ can be formulated as designing the Bayesian inference experiments 
 which  maximizes the  expected utility $\mathcal{U}$ in  the design space $\mathbb{R}^{d\times N_D}$:
\begin{equation}
\label{eq:optimalD}
D^* = \mathop{\arg\max}_{D\in \mathbb{R}^{d*N_D}} \mathcal{U}(D),
\end{equation}
where $D$ is the set of $N_D$ locations in $\mathbb{R}^d$. The  practical computation is an approximation  of this utility function  based on
the current learned machine learning model, not the true original model.
Following~\cite{huan2013simulation,wang2016gaussian}, 
we introduce 
the following  {\it expected utility function} 
(also named as {\it  the acquisition function})

\begin{equation}
\label{eq:U0}
\begin{split}
\mathcal{U}(D) &= \iint \omega(D, \y, {Z})p( {Z}, {\y}| D)d {Z} d {\y} = 
 \iint \omega(D, {\y}, {Z}) p( {Z} |  \y, D) p( {\y} | D) ~d {Z} d {\y},
\end{split}
\end{equation}

where  $\omega(D,\y,  {Z})$ is a utility function to be specified in practice.
$p(\y\vert D)$ is the distribution of the prediction of the current surrogate model  at the locations $D$.
$p({Z} \vert {\y},D)$ is the distribution  of the inferred parameter ${Z}$
of the updated model after retrained  by 
  assigning the  labels ${\y}$ to  the new design locations $D$. We call $p({Z} \vert {\y},D)$
  the posterior distribution of $Z$ and  denote the  prior distribution based on the current surrogate model by $p(Z)$. 
 

 A popular choice of the utility function is the  {Kullback-Leibler divergence (KLD)}, also known as the relative entropy, 
 between the posterior and the prior distributions ~\cite{wang2016gaussian}.
  For two distributions $p_A(Z)$ and $p_B(Z)$, the KLD from $p_A$ to $p_B$ is defined as 
  \begin{equation}\label{criteria}
 \operatorname{D_{KL}}(p_A||p_B) = \int p_A(Z)\log [\frac{p_A(Z)}{p_B(Z)}] dZ = \mathbb{E}_A[\log \frac{p_A(Z)}{p_B(Z)}],
 \end{equation}
where we define $0\log0 \equiv 0$. Equation \eqref{criteria} often serves as a criterion for measuring the difference between two distributions. Then,
the utility function $\omega$ is independent of $Z$:
 \begin{equation} \label{624}
\omega_{KL}(D,\y,Z) =\omega_{KL}(D,\y) \triangleq  \operatorname{D_{KL}}(p(\cdot|\y, D) || p(\cdot)) = \int p(Z|\y, D)\log [\frac{p(Z|\y, D)}{p(Z)}] dZ.
\end{equation} 
 The utility function $\omega_{KL}(D,\y)$ can be understood as the information gained by performing experiments under the conditions $D$, and a  larger value of $\omega_{KL}$ implies that the experiment is more informative for  inference.
Substitute this $\omega_{KL}$ into  \eqref{eq:U0}, then we obtain 
 \begin{align}
 \label{eq:U}
\mathcal{U}(D) &= \iint \omega_{KL}(D,\y)  p(Z,\y|D)dZ d\y  =  \int \omega_{KL}(D,\y)  p(\y |D)  d\y  \nonumber \\
                    & = \iint p(Z,\y|D)\log \frac{p(Z,\y|D)}{p(Z)p(\y|D)} dZ d\y.
\end{align}
We can see that the expected utility function \eqref{eq:U} is just the  {\it mutual information} between the parameter $Z$ and the data $\y$.
Indeed, an equivalent interpretation of Equation \eqref{eq:U} is to select $\omega(D, \y, {Z})=\log \frac{p(Z,\y|D)}{p(Z)p(\y|D)}$ in Equation \eqref{eq:U0}.
This choice of $\mathcal{U}$ is thus well justified since the optimal choice of $D^*$  can offer the most mutual information of $p(Z,\y)$
so that if the new design and observations  $(D^{*},\y)$ are used to retain the model, the  inference $Z$ benefits most from this information of $\y$.
On the other hand,  the worse case of $\mathcal{U}(D)=0$ implies that for such designs $D$, the observed data $\y$ has zero contributions to 
infer the parameter $Z$. The use of \eqref{eq:U} can be traced back to \cite{huan2013simulation} and references therein.

The form of the expected utility function \eqref{eq:U} has to be approximated numerically 
and it requires  to evaluate and sample $Z$ and $\y$ for each $D$,  which constitutes
the  significant computational cost.  In the following two subsections, we discuss how to simplify 
the formulation and implement the algorithm for  GAD.

\subsection{Sub-optimal design criterion }
Here we propose a sample-based sub-optimal design criterion.
 We rewrite  equation \eqref{eq:U} as the difference 
\begin{align}
\label{eq:U_entropy}
\mathcal{U}(D) &= \ \int \omega_{KL}(D,\y)  p(\y |D)  d\y  \nonumber \\
          &=  \iint p(Z|\y,D)\log p(Z|\y,D) d Z p(\y|D)d\y - \iint p(Z|\y,D) p(\y|D)d\y \log p(Z) d Z \nonumber \\
          &=  \iint p(Z|\y,D)\log p(Z|\y,D) d Z p(\y|D)d\y - \int p(Z|D) \log p(Z) d Z \nonumber  \\
          & := \mathcal{U}_1(D) - \mathcal{U}_2(D).
\end{align}  
Since $p(Z\vert D)$ conditioned on the locations $D$ only  has integrated out the output labels $\y$,  
we have $p(Z|D) = p(Z)$ and then $\mathcal{U}_2$ is a constant: 
$ \mathcal{U}_2(D) = \int p(Z|D) \log p(Z) d Z = \int p(Z) \log p(Z) d Z  = const$.
For $\mathcal{U}_{1}$, we have
\begin{align}
\label{675}
 \mathcal{U}_1(D) &= \int \left( \int p(Z|\y,D)\log p(Z|\y,D) d Z \right) ~p(\y|D)d\y= \int -\H\left(p(Z|\y,D\right) )~p(\y|D)~d\y,
 \end{align}
where $$ \H\left(p(Z|\y,D)\right)= -\int p(z|\y,D)\log p(z|\y,D)~dz$$ is the entropy of the posterior distribution $p(Z|\y,D)$. 
 The intuition
behind this expression is that a large value $\mathcal{U}_{1}(D)$   implies
that the data $\y$ at $D$ decreases the entropy in $Z$, and hence those data are
more informative to infer $Z$.

Note that $p(Z\vert \y, D)$ is the posterior distribution of  $Z$ 
of the updated model after retrained  by 
  assigning the  labels ${\y}$ to  the new design locations $D$. 
  In general, it could be a very complicated distribution.
But in case of  the Gaussian distribution $p(Z|\y,D)=\mathcal{N}(Z;\mu,\Sigma)$, we have  the entropy expression $\H(p(Z|\y,D))= \frac{1}{2}\log((2\pi e)^{d} \det(\Sigma))$, where $e$ is the Euler's number.
%
%
\subsection{GAD sample-based criterion }
We recall that in our main method specified in Section \ref{main_result},
the position variable $\x$ marches from $\x^{(t-1)}$ to $\x^{(t)}$ and the direction variable $\v$
updates from $\v^{(t-1)}$ to $\v^{(t)}$ at each time step indexed by $t$, by the  GAD  \eqref{HGADv} 
whose coefficient functions $\b$ and $J$ are based on the   surrogate model.   
This surrogate model  will be retrained, if it is deemed to have insufficient fidelity, 
by combining the existing dataset and a set of $N_D$ new observations at the design locations $D^{*}$.
To determine $D^{*}$ by the  above active learning strategy,  we choose the ``parameter'' of interest $Z$ as  the future trajectory of the GAD by using Equation \eqref{HGADv}   starting from  the current 
point $\x^{(t-1)}$.  By placing special emphasis on  this GAD's future trajectory rather than pursuing  the overall 
quality of the surrogate model on the whole space,   our specific  active learning method here can provide the 
set of  “next-best” points  for our underlying dynamics of interests. 
Below are the details on the criterion \eqref{675} used in our active learning.

We need to specify the choice of $Z$ and its posterior distribution $p(Z\vert \y,D)$
for any given pair of $(\y, D)$.
We choose $Z$ as  the continuous-time GAD path up to a certain time $T$,  $Z=\{\z(t'): t'\in [0,T] \}$, 
which satisfies the GAD \eqref{HGADv}  associated with the  functions  $\tilde{\b}$ and $\tilde{J}$ sampled from 
 the {\it posterior} distribution 
obtained after retrainng the GPR with the addition of any pair of the given data $(\y, D)$:
\begin{subequations}
\begin{empheq}[left=\empheqlbrace]{align}
 \dot{\z}(t')  &=  \tilde{\b}(\z)-2 \inpd{\tilde{\b}(\z)}{\v(t')} \v(t') / \norm{\v(t')}^2,
 \\
   \dot{\v}(t') &=  \tilde{J} (\x) \v - \inpd{\v}{\tilde{J} (\x) \v} \v,   \end{empheq}
\end{subequations}
with the initial $ \z(t'=0)= \x^{(t)}, ~\v(t'=0)=\v^{(t)}$.
 To ease the presentation, we use its discrete-time representation $Z=(\z^{1},\ldots, \z^{k},\ldots)$, $\z^{k}\in \Real^{d}$.
If the forward Euler scheme is used, we have  for $1\leq k\leq K$,
\begin{subequations}{\label{HGADv_dis2}}
\begin{empheq}[left=\empheqlbrace]{align}
 \z^{k}  &= \z^{k-1} + \Delta t ~
 \left( \tilde{\b}(\z^{k-1})-2 \inpd{\tilde{\b}(\z^{k-1})}{\v^{k-1}} \v^{k-1} / \norm{\v^{k-1}}^2\right),
 \\
   \v^{k}  &=\v^{k-1} + \Delta t ~\left(  \tilde{J} (\z^{k-1}) \v^{k-1} - \inpd{\v^{k-1}}{\tilde{J}(\z^{k-1})  \v^{k-1} } \v^{k-1} \right),
 \end{empheq}
\end{subequations}
where $\Delta t$ is the time step size (which could be different from the time step size in \eqref{HGADv_dis}), $\z^{0}=\x^{(t)}$ and $\v^{0}=\v^{(t)}$.
To have a tractable form of the entropy of 
the distribution of $(\z^{1},\ldots,\z^{K})$, we shall make the following  two important approximations.

The first is to approximate the iteration  \eqref{HGADv_dis2} for the pair $(\z^k, \v^k)$ by 
the following iteration involving $\z^k$ only
\begin{equation}
\label{eq:linearization}
\z^{k} \approx \z^{k-1} + \Delta t \left( \alpha \tilde{\b}(\z^{k-1}) + \bm{\beta} \tilde{J}(\z^{k-1}) \right),
\end{equation}
where the scalar $\alpha$ and  the vector $\bm{\beta} = [\beta_1,\dots,\beta_d]$ are the  hyper-parameters  which will be determined later by the least square method. This approximation makes sense because the dynamics $\z$ in the GAD
is essentially driven by the force field  and the Jacobi matrix.
The approximation \eqref{eq:linearization} immediately implies that $(\z^{k})$ 
now is  a discrete-time Markov process whose  transition probability  
 is a Gaussian distribution  determined by
 the posterior distribution of  the $\mathbb{R}^{(d+d^{2})}$-valued Gaussian random function $(\tilde{\b}   ,\tilde{J})$ at the 
 position $\z^{k-1}$ (see \eqref{eq:postCov}).
In what follows  we assume $Z=(\z^k)$ satisfies this approximation and  focus on the derivation of  the analytical form of the entropy of
the posterior distribution $p(Z\vert \y,D)$, which will be used in the utility function \eqref{675}.
By the chain rule of the entropy \cite{2006Elements}, the entropy  of the path distribution $p(Z\vert \y, D)=p(\z^1,\ldots, \z^K \vert \y, D)$ can be written as the entropy of the transition probability $p(\z^k \vert  \z^{k-1}, \y ,D)$: 
\begin{equation}
\label{eq:sumH}
 \H(p(Z\vert \y, D))=\sum_{k=1}^K \e_{\z^{k-1}}\H(p( \z^k\vert  \z^{k-1},  \y, D)). 
 \end{equation}
 If  we explicitly denote   the covariance of the Gaussian transition probability $p( \z^k\vert  \z^{k-1},  \y, D)$
 as $\tilde{\Sigma}_{post}(\z^{k-1} ;\y ,D)$, then
\begin{equation}
\label{784}
 \H(p(Z\vert \y, D))=\sum_{k=1}^K \e_{\z^{k-1}} \frac12 \log \left((2\pi e)^d \det\left(\tilde{\Sigma}_{post}(\z^{k-1} ;\y ,D)\right) \right). 
 \end{equation}
The expectation $\e_{\z^{k-1}}$ in \eqref{784} is the marginal distribution of  the {\it posterior path distribution} $p(Z\vert \y, D)$. 
Here $\tilde{\Sigma}_{post}(\z^{k-1} ;\y ,D)$ emphasises 
 the dependence on   $\z^{k-1}$, $\y $ and $D$.
 However, this covariance $\tilde{\Sigma}_{post}(\z^{k-1} ;\y ,D)$ in fact
 is independent of the new labelled data $\y$ because 
 the posterior covariance  of the GPR 
 does not involve the labels  $\y$  (see \eqref{eq:postCov}).
 Therefore $\tilde{\Sigma}_{post}(\z^{k-1} ;\y ,D)$=$\tilde{\Sigma}_{post}(\z^{k-1} ;D)$
 and the utility function 
 $\mathcal{U}_1$  in  \eqref{675}
is simplified as 
\begin{align} \label{802}
 \mathcal{U}_1(D)  &=   \int-\H\left(p(Z|\y,D\right) )~p(\y|D) d\y =
- \sum_{k=1}^K \e_{\z^{k-1}} \frac12 \log \left((2\pi e)^d \det\left(\tilde{\Sigma}_{post}(\z^{k-1} ; D)\right) \right). 
 \end{align}
 
To further reduce the cost
from retraining the  force field and the Jacobi matrix
for each $\y$ and $D$, we 
propose the second approximation here, which is to replace
the posterior   $\e_{\z^{k-1}}$ in \eqref{eq:sumH} and \eqref{784} by their prior  
version  since sampling the prior GAD path does not involve $\y$ and $ D$ at all.
Specifically, this means that we simulate a sequence $\widehat{\z}^k$ by the  GAD
 where the force field and the Jacobi matrix, denoted by $\widehat{b}$ and $\widehat{J}$  respectively, 
are sampled from the prior distribution of the current  Gaussian surrogate model
\begin{subequations}{\label{HGADv_dis3}}
\begin{empheq}[left=\empheqlbrace]{align}
\widehat{ \z}^{k}  &= \widehat{ \z}^{k-1} + \Delta t ~
 \left( \widehat{\b}(\z^{k-1})-2 \inpd{\widehat{\b}(\widehat{\z}^{k-1})}{\widehat{\v}^{k-1}} \widehat{\v}^{k-1} /
  \norm{\widehat{\v}^{k-1}}^2\right),
 \\
  \widehat{ \v}^{k}  &=\widehat{\v}^{k-1} + \Delta t ~\left(  \widehat{J} (\widehat{\z}^{k-1})
   \widehat{\v}^{k-1} - \inpd{\widehat{\v}^{k-1}}{\widehat{J}(\widehat{\z}^{k-1}) \widehat{ \v}^{k-1} } \widehat{\v}^{k-1} \right),
\end{empheq}
\end{subequations}
with initial $\widehat{ \z}^{0}=\x^{(t)} $ and $\widehat{ \v}^{k} =\v^{(t)}$.
Consequently, the utility function  in \eqref{802}  becomes
\begin{align}
\label{eq:U18}
 \mathcal{U}_1(D)  &\approx  
 -\sum_{k=1}^K \e_{\widehat{\z}^{k-1}} \frac12 \log \left((2\pi e)^d \det\left(\tilde{\Sigma}_{post}(\widehat{\z}^{k-1} ; D)\right) \right),
 \end{align}
 where   $\tilde{\Sigma}_{post}(\widehat{\z}^{k-1} ; D)$
 is the covariance matrix of the Gaussian distribution of  
\begin{equation}
\label{eq:linearization2}
\z^{k} := \widehat{\z}^{k-1} + \Delta t \left( \alpha \tilde{\b}(\widehat{\z}^{k-1}) + \bm{\beta} \tilde{J}(\widehat{\z}^{k-1}) \right).
\end{equation}

 \subsection{Implementation}
 
In the last part of this section, we shall present further  details and discuss how to maximize the utility function.
The main task of choosing the optimal  $N_D$ design locations $D^*$ is now 
the maximization problem of $\mathcal{U}_1(D)$ defined in \eqref{eq:U18}. 
The expectation $\e_{\widehat{\z}^{k-1}} $ in 
 \eqref{eq:U18} is approximated by the sample average
 by sampling a few realization of the Gaussian surrogate model $\widehat{\b}$ and $\widehat{J}$.
Following the recommendation of \cite{huan2013simulation}, we use the simultaneous perturbation stochastic approximation (SPSA) method  \cite{spall1992multivariate, spall1998implementation} to solve the optimization problem \eqref{eq:optimalD}.
SPSA is a derivative-free stochastic optimization method and we provide the detailed algorithm of SPSA in Appendix \ref{appspsa}.

For practical efficiency, we develop  some other details of approximations
 to speed up the numerical implementation. It is worthwhile to point out  that
 the pursue of the high accuracy of $\mathcal{U}_1(D)$ is not 
necessary here and the trade-off between the precision and the efficiency
for computation of $\mathcal{U}_1(D)$ is necessary.

\begin{itemize}
\item {Approximate calculation of the posterior covariance of $Z$.}

  $\det\left(\tilde{\Sigma}_{post}(\widehat{\z}^{k-1} ; D)\right)$ in \eqref{eq:U18}
can be approximated  by using  the product of its diagonal elements only
to make it tractable in practice. This implies the assumption that the  $d$ components
of  the GAD path $\widehat{\z}$ has no correlation.
We drop off  the argument $D$ for easy notation and write  $\widehat{\z}^{k-1}$ as a generic
variable $\z$ in $\tilde{\Sigma}_{post}(\widehat{\z}^{k-1} ; D)$, so this matrix 
is referred to as   $\tilde{\Sigma}_{post}(\z)$.
By the definition \eqref{eq:linearization2}, 
the diagonal elements are 
$$\alpha^2 \var(\tilde{b}_i(\z))+ \sum_{j=1}^d \beta_j^2 \var(\tilde{J}_{j,i}(\z)) + 2 \alpha \beta_j \Cov(\tilde{b}_i(\z),\tilde{J}_{ji}(\z)).
$$
The posterior variances above are given by the result in \eqref{eq:postCov},
for example,  
 \begin{equation}
\begin{split}
\var(\tilde{b}_i(\z)) &= K_{b_i,b_i}(\z,\z)-K_{b_i,u}(\z,\X^*\cup D)K_{u,u}(\X^*\cup D,\X^*\cup D)^{-1}K_{u,b_i}(\X^*\cup D,\z), \\
\var(\tilde{J}_{i,j}(\z))&= K_{J_{i,j},J_{i,j}}(\z,\z)-K_{J_{i,j},u}(\z,\X^*\cup D)K_{u,u}(\X^*\cup D,\X^*\cup D)^{-1}K_{u,J_{i,j}}(\X^*\cup D,\z),
\end{split}
\end{equation} 
where $\X^*$ is the location of the train dataset in history used for the current surrogate model
and $D$ is the new design location. 
The size of   $\X^*\cup D$ grows by $N_D$ at each iteration of model update.
When this size is too large, we use the new design point $D$ only  to
keep the computation complexity of the  posterior variance low at a constant $\mathcal{O}((N_D)^3)$
instead of growing with the size of the total dataset $\X^*\cup D$.
Thus, we have
 \begin{equation}
\label{eq:variance_fast}
\begin{split}
\var(\tilde{b}_i(\z))  &\approx K_{b_i,b_i}(\z,\z)-K_{b_i,u}(\z,D)K_{u,u}(D,D)^{-1}K_{u_i,b}(D,\z), \\
\var(\tilde{J}_{i,j}(\z)) &\approx K_{J_{i,j},J_{i,j}}(\z,\z)-K_{J_{i,j},u}(\z,D)K_{u,u}(D,D)^{-1}K_{u,J_{i,j}}(D,\z).
\end{split}
\end{equation} 
This approximation is reasonable because 
the sample path $\z$ of interest starts from  the current  position $\x^{(t-1)}$
and the  previous  locations in  $\X^*$ may not be as relevant as the new  design locations $D$.

\item {Determination of $\alpha$ and $\boldsymbol{\beta}$ by the least square fitting}

The reduction in \eqref{eq:linearization} is important to derive the distribution of $\z^k$, which  requires  the determination of two
hyper-parameters $\alpha$ and $\boldsymbol{\beta}$.
These hyper-parameters can be computed easily by the least square method with $2d$ equations 
\begin{equation}
\left\{
\begin{split}
\frac{\x^{(t)} - \x^{(t-1)}}{\Delta t} &=  \alpha \bm{\mu}^b(\x^{(t-1)}|\y,D) + \bm{\beta} \bm{\mu}^J(\x^{(t-1)}|\y,D), \\
\frac{\x^{(t-1)} - \x^{(t-2)}}{\Delta t} &=  \alpha \bm{\mu}^b(\x^{(t-2)}|\y,D) + \bm{\beta} \bm{\mu}^J(\x^{(t-2)}|\y,D),
\end{split}
\right.
\end{equation}
and stored variables $\x^t$, $\x^{t-1}$,$\x^{t-2}$, $\bm{\mu}^b(\x^{t-1}|\y,D)$, $\bm{\mu}^b(\x^{t-2}|\y,D)$, $\bm{\mu}^J(\x^{t-1}|\y,D)$ and $\bm{\mu}^J(\x^{t-2}|\y,D)$.
\end{itemize}

In summary, Algorithm \ref{alg:optimal_design} summarizes the details of computing optimal design $D^*$.

\IncMargin{1em}
\begin{algorithm}[H]\label{alg:optimal_design}
\caption{Compute the optimal design $D^*$}
\LinesNumbered 
\SetKwInOut{Input}{Parameter}\SetKwInOut{Output}{Output}
\SetAlgoLined
\Input{Number of locations $N_D$; Gaussian surrogate model $\mathcal{F}$;  Numbers of paths $n$; Max time $T$; time step $\Delta t$;
 location $\x $ and its corresponding direction $\v$;}
\Output{Optimal designs $D^*=(\d_{1},\ldots, \d_{N_{D}})$; }
 $//$ Sample $n$ path  $Z_1,\dots,Z_n$ from prior $p(Z)$ \;
Set $K=T/{\Delta t}$, initial condition $\widehat{\z}^{0}=\x$ and $\widehat{\bm{v}}^{0}=\bm{v}$\;
\For{$i=1$ \KwTo $n$} 
{Get the $i$th sample functions denoted as $\hat{\b}_i$ and $\hat{J}_i$ which are sampled from Gaussian processes $ \mathcal{F}$\;
\For {$k=1$ \KwTo $K$}{
Compute $Z_k = \{ \widehat{\z}^1,\dots,\widehat{\z}^K \}$ by Equation \eqref{HGADv_dis3} with $\Delta t$, the sample functions $\hat{\b}_i$ and $\hat{J}_i$\;
}
}
Solve the maximization problem  $D^* =  \mathop{\arg\max}_{D\in \mathbb{R}^{d*N_D}} \mathcal{U}_{1}(D)$ by SPSA algorithm 
where $\mathcal{U}_{1}$ is defined in Equation \eqref{eq:U18}\ ;
\end{algorithm}
\DecMargin{1em}
The exploration time interval $T$ or  the corresponding steps $K$ is in general not  necessarily very large, for instance $K=10$ is enough.

%

\section{Numerical examples}\label{ex}
In this section we first consider the SP detection problem in two mathematical examples  so that we can validate the results with exact solutions.
The first example a gradient system with a known energy function. The second example in Section \ref{set:non_gradient} shows 
how  to handle non-gradient systems and  the robustness of the method with respect to the noisy observation.
In Section \ref{set:alanine}, we apply our method to a computational intensive MD simulation problem: the alanine dipeptide model, which is a benchmark  problem tested in literature \cite{li2019computing,pan2008finding,Maragliano2006StringMI}. 

\subsection{An example of gradient system}\label{set:gradient}
We consider a gradient system whose energy function is 
\begin{equation} 
\label{eq:example1}
u(\bm{x}) = \frac12 \bm{x}^\mathrm{T} M\bm{x} - 5 R(\bm{x}),
\end{equation}
where $M$ = $\begin{bmatrix}
0.8 &-0.2\\
-0.2 &0.5
\end{bmatrix}$ and  $R(\bm{x}) = \sum^2_{i=1} \arctan (x_i - 5)$. 
To facilitate the comparison of the classical GAD and the aGPR-GAD methods, in this example we consider the case where the observation data has no noise pollution: $y_i = u(\x_i)$. 
For this gradient system, there are three local minimums $m_1=(0.46,0.69)$, $m_2=(2.20,5.98)$, $m_3=(5.71,6.23)$ and two saddle points $s_1 =(1.28,3.44)$, $s_2 =(3.56,6.07)$ in the domain $[-1, 7]\times[-1, 7]$.

In our aGPR-GAD method, the threshold $\sigma^2_{sur}=0.2$ in Algorithm\ref{alg:gad-gpr1} and the step size $\Delta t =0.01$ in both Algorithm\ref{alg:gad-gpr1} and Algorithm\ref{alg:optimal_design}.
 $N_{D}=10$, $T = 0.1$  and   $n=20$ in Algorithm\ref{alg:optimal_design}.
We draw $20$ initial data locations by a dimensional-independent Gaussian distribution whose mean is  $m_1$ or $m_2$ and variance is $0.5$ for each dimension. 
In   the classical GAD method, we choose   $\Delta t  = 0.1$.

\begin{figure}
  \centering
	\includegraphics[width=0.85\linewidth]{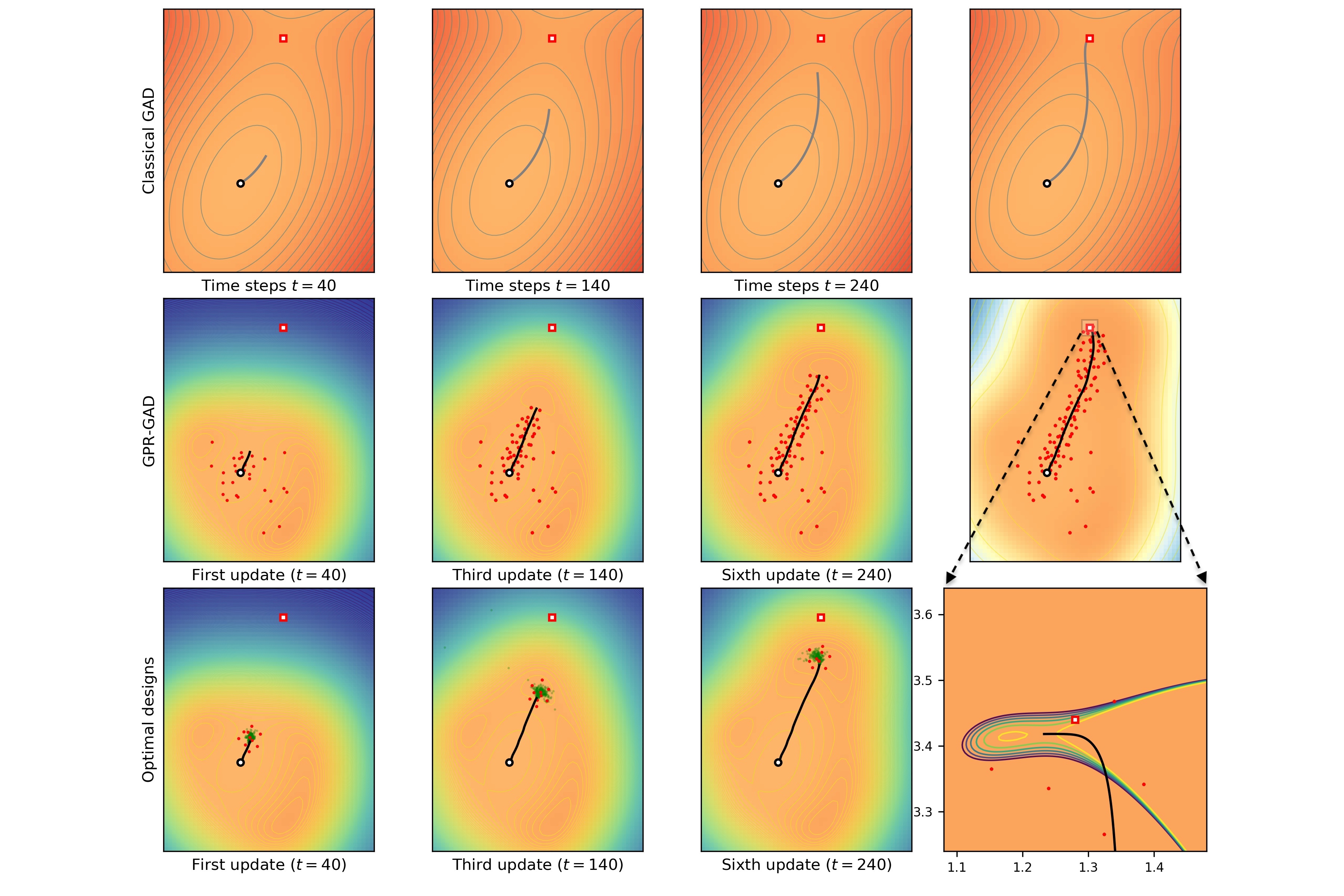}
	\caption{Comparison between the classical GAD   and aGPR-GAD  methods. 
	The local minimum $m_1$ is marked by white circle with black edge and the saddle point $s_1$ is marked by white square with red edge. 
	{\it Top Row:}  The trajectories (gray curves) computed by classical GAD method up to  time steps $t=40,140, 240$ and the final convergent time respectively.
	The contour is     the true energy surface $u$ in \eqref{eq:example1}.
	{\it Middle Row:} The trajectories (black curves) computed by aGPR-GAD method and the positions of the train data (red stars) at the first GPR update ($t=40$),  third GPR update ($t=140$) and sixth  GPR update ($t=240$), and ninth GPR update (the last  convergent time). 
	The contours are the corresponding  learned surrogate energy functions.
	 The  small area near $s_{1}$ in the last subfigure is zoomed in view at the bottom row.
	{\it Bottom Row:}  The left three subfigures show the  batch of new optimal design points $D^*$ (red dots) selected by active learning to update GPR.
	The green dots indicate the prior sample $\hat{\z}^k$ used to estimate $\mathcal{U}_{1}$}
		\label{f:eg1_1}
\end{figure}

We first illustrate the SP detection procedure starting from $m_1$ in detail.
The subfigures in Figure \ref{f:eg1_1} on the first two rows show the trajectories at different time steps computed by classical GAD and aGPR-GAD, respectively.
Both methods tend to  find the approximation to the SP $s_1$, but aGPR-GAD has a lower accuracy; see Table \ref{tab:example1} below.
The aGPR-GAD takes nine active learning updates of selecting new designing points ( i.e., calling  Algorithm\ref{alg:optimal_design} nine times).
The panels in the second row show the aGPR-GAD trajectory and the datasets(red dots) associated with the current GPR model $\mathcal{F}$, up to the moment where the surrogate model $\mathcal{F}$ is determined as unreliable  (Line 8 in Algorithm\ref{alg:gad-gpr1}); and then  the {\it new} data points are added by calling  Algorithm\ref{alg:optimal_design}, as shown in the first three panels in the last row of Figure \ref{f:eg1_1}. It is observed that  the increased design points are indeed  around the trajectory toward the SP and it shows our active learning Algorithm\ref{alg:optimal_design}
can effectively guide the sample points for query towards the SP.

\begin{figure}
  \centering
	\includegraphics[width=0.45\linewidth]{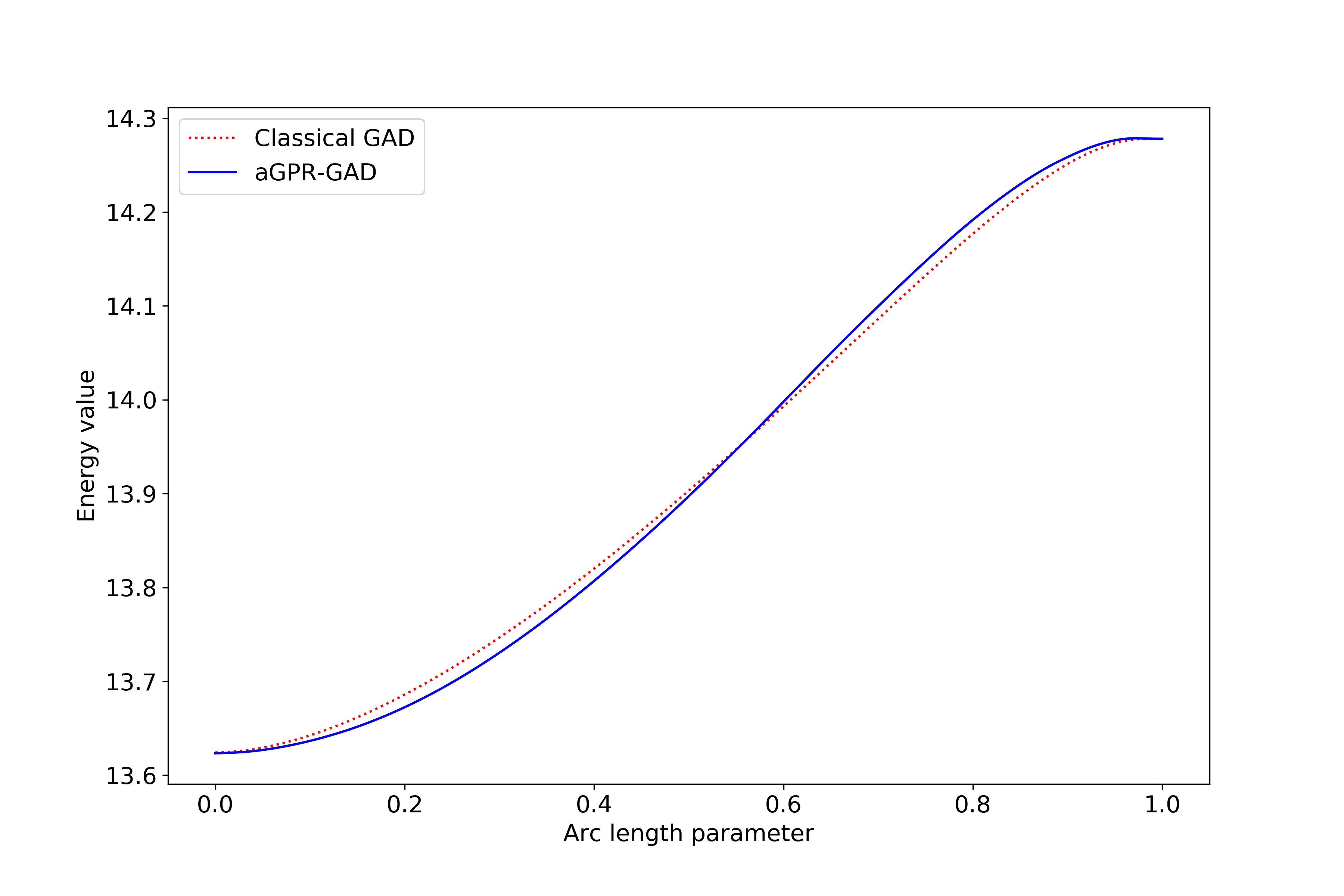}
	\caption{The energy function value with respect to arc length parameter of the trajectories computed by the classical GAD and the aGPR-GAD, respectively.} 
	\label{f:eg1_5}
\end{figure}

The reader can notice  that the contour plots of the true energy potential  \eqref{eq:example1} (the first row) and the surrogate energy function 
(the mean of the GPR)  are quite different, even when the aGPR-GAD successfully finds the approximate SP.
Indeed, since our design points concentrate on a local tubular region along the GAD trajectory, the surrogate model can   deviate the true energy function significantly 
away from the trajectory.  The magnified contour of the surrogate energy function near the SP   in the last panel of Figure  \ref{f:eg1_1} even shows the local topology of saddle point 
may be also different.  However,  what matters here is the good approximate of the energy function around the GAD trajectory. 
We plot the true and surrogate energy functions along the arc-length  parametrized curves of GAD trajectories for the classic GAD and the aGPR-GAD, respectively, in Figure \ref{f:eg1_5}. 
This result shows that the design points selected by our method  truly reflect the information for  the GAD trajectory.

\begin{figure}[htbp]
  \centering
 	\includegraphics[width=.45\linewidth]{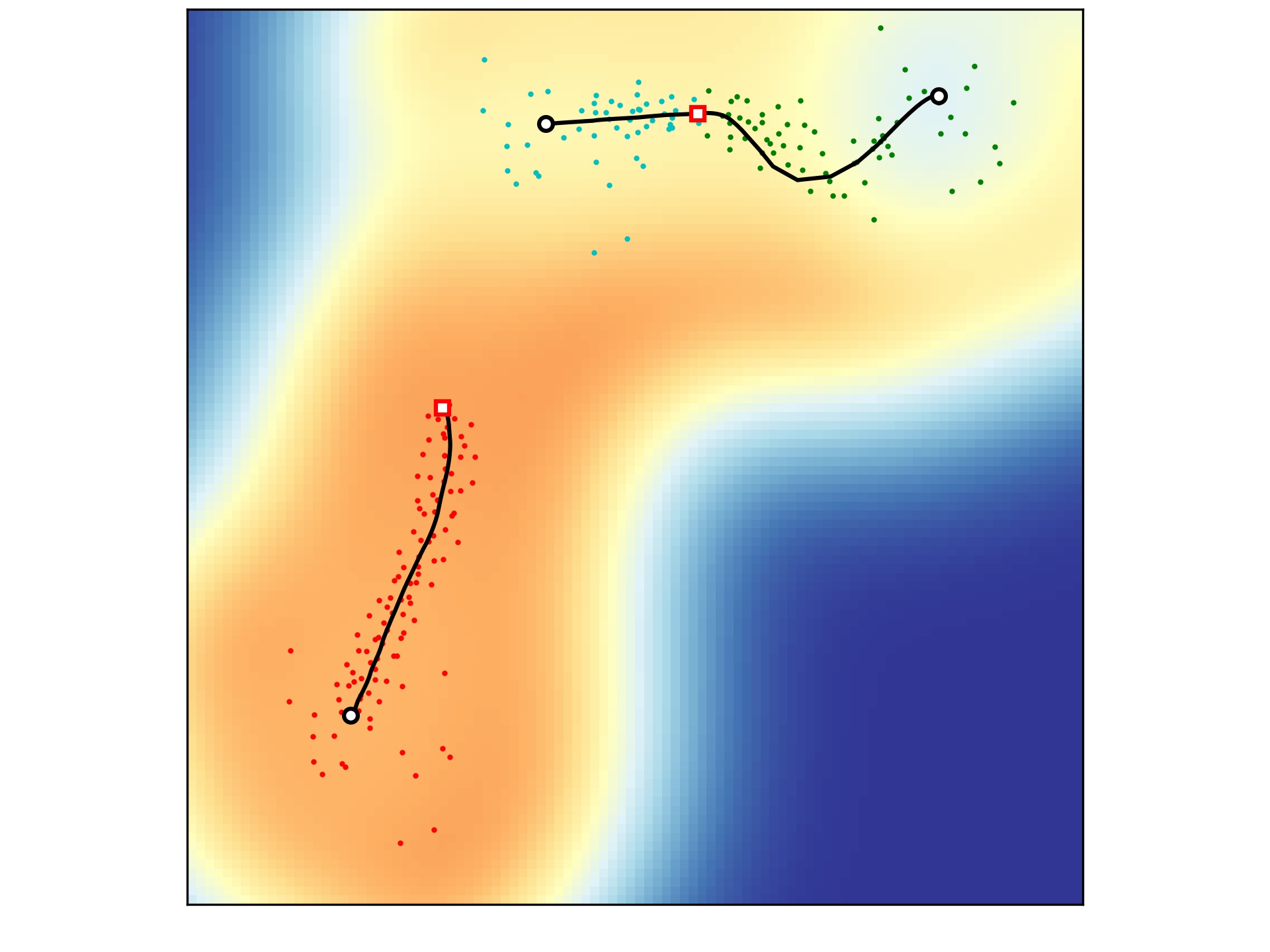}
	\caption{The convergent trajectories  computed by aGPR-GAD starting from local minima  $m_1$, $m_2$ and $m_3$, respectively.
	The  colored dots  represent the data locations in total sampled by the active learning  and the background image shows the landscape of the surrogate function
	(trained in final state by using all data points for three trajectories).
	The local minimums are marked by white circle with black edge and the saddle points are marked by white square with red edge.} 
	\label{f:eg1_2}
\end{figure}

Then we test our aGPR-GAD method by starting from three different local minimums ($m_1$, $m_2$ and $m_3$, respectively) and three trajectories are presented in Figure \ref{f:eg1_2}.
The results that all the end points of trajectories are next to SP and the design locations are around the trajectories. 
 Table \ref{tab:example1} present their numerical values.  Their accuracy could  be improved  if  one reduces the threshold  $\sigma^2_{sur}$.

To consider  the computing cost,  we list a number $N^{*}$ in Table \ref{tab:example1},
which is related to the cost but has different meanings for the classical GAD and aGPR-GAD methods. 
 $N^{*}$ indicates the total number of queries of the function $u$ in aGPR-GAD method, i.e.,  $N^{*} = |\mathcal{D}|$ the total number of design points, and in classical GAD method it indicates the number of discrete points 
 $\set{\x^{(t)}}$ consisting of the  trajectory (so, $N^{*}$ calls to the gradient and $N^{*}$ calls to the Hessian).
Note  that the step size in the classic GAD is ten times larger than the one used in the aGPR-GAD.
We can find $N^*$ in aGPR-GAD method are much less than the one in classical GAD with same starting point, which means less computing cost of the aGPR-GAD method are used.

\begin{table}[htbp]
{\footnotesize
  \caption{ Specification and results for the gradient system example.}  
  \label{tab:example1}
\begin{center}
  \begin{tabular}{|c|c|c|c|} \hline
     initial point $\x^{(0)}$ & method & $\x^{SP}$ & $N^{*}$ \\ \hline
     $(0.46, 0.69)$   & GAD & $(\bm{1.28,3.44})$ & $305$ \\
       & aGPR-GAD & $(1.15, 3.36)$ &  $110$\\\hline
      $(2.20, 5.98)$ & GAD & $(\bm{3.56,6.07})$ & $178$\\
       & aGPR-GAD & $(3.53,6.06)$ & $50$\\\hline
     $(5.71, 6.23)$  & GAD & $(\bm{3.56,6.07})$ & $398$\\
       & aGPR-GAD & $(3.62,6.07)$ &  $70$\\\hline
  \end{tabular}
\end{center}
}
\end{table}

\subsection{Two dimension non-gradient system}
\label{set:non_gradient}
Our second example is the following two dimensional non-gradient system
\begin{align}
\label{eq:nongradient}
\bm{b}(\x)= - \frac{1}{2} \sum_{j=1}^2 A_{ij}x_{j} + 5\Gamma_i(\bm{x}),i=1,2,
\end{align}
where $A = \begin{bmatrix}
0.8 &-0.3\\
-0.2 &0.5
\end{bmatrix} $, $\Gamma_i(\bm{x}) = [1 + (x_i-5)^2]^{-1}$.
This dynamics has two stable fixed points $m_1=(0.59, 0.76)$, $m_2 = (5.87, 6.25)$ and a unique saddle point $s = (1.79, 3.30)$.

We use this  example to illustrates the ability of our aGPR-GAD method when the measurement is the  vector value of the force field,
in particular,  the measurement data is  noisy.  The measurement data are generated as the following  noisy   vector  field
\begin{equation}
\label{eq:example2}
\y = \bm{b}(\x) + \bm{\epsilon},
\end{equation}
where $\bm{\epsilon}$ is   Gaussian   noise with the  variance $\sigma^2_{u}I_{2\times 2}$.

We set $\Delta t =0.01$ in both Algorithm \ref{alg:gad-gpr1} and Algorithm \ref{alg:optimal_design}, $T = 0.1$, $n=20$ and $N_D=10$ in Algorithm \ref{alg:optimal_design}.
  20 initial data locations are generated  from
the Gaussian distribution centred at   $m_1$ or $m_2$ with  variance $0.3I_{2\times2}$.
The initial data-set is composed by these initial data locations and their corresponding force field values  by equations \eqref{eq:example2}.
Since now the observation is a two-dimensional force field, we construct two independent GPR surrogates for each component of $\b(\x)$ in this problem. 
Then the estimation of Jacobi matrix $D \b(\x)$ can be formulated as the first order derivative estimation with the given force field observation  by using GPR in Section \ref{set:GPR}.
 In classical GAD method, we set $\Delta t = 0.01$ and 
 the force field is also the noisy observation defined in \eqref{eq:example2} and the Jacobi matrix is computed by finite difference.
  
\begin{figure}[htb]
  \centering
	\includegraphics[width=.61\linewidth]{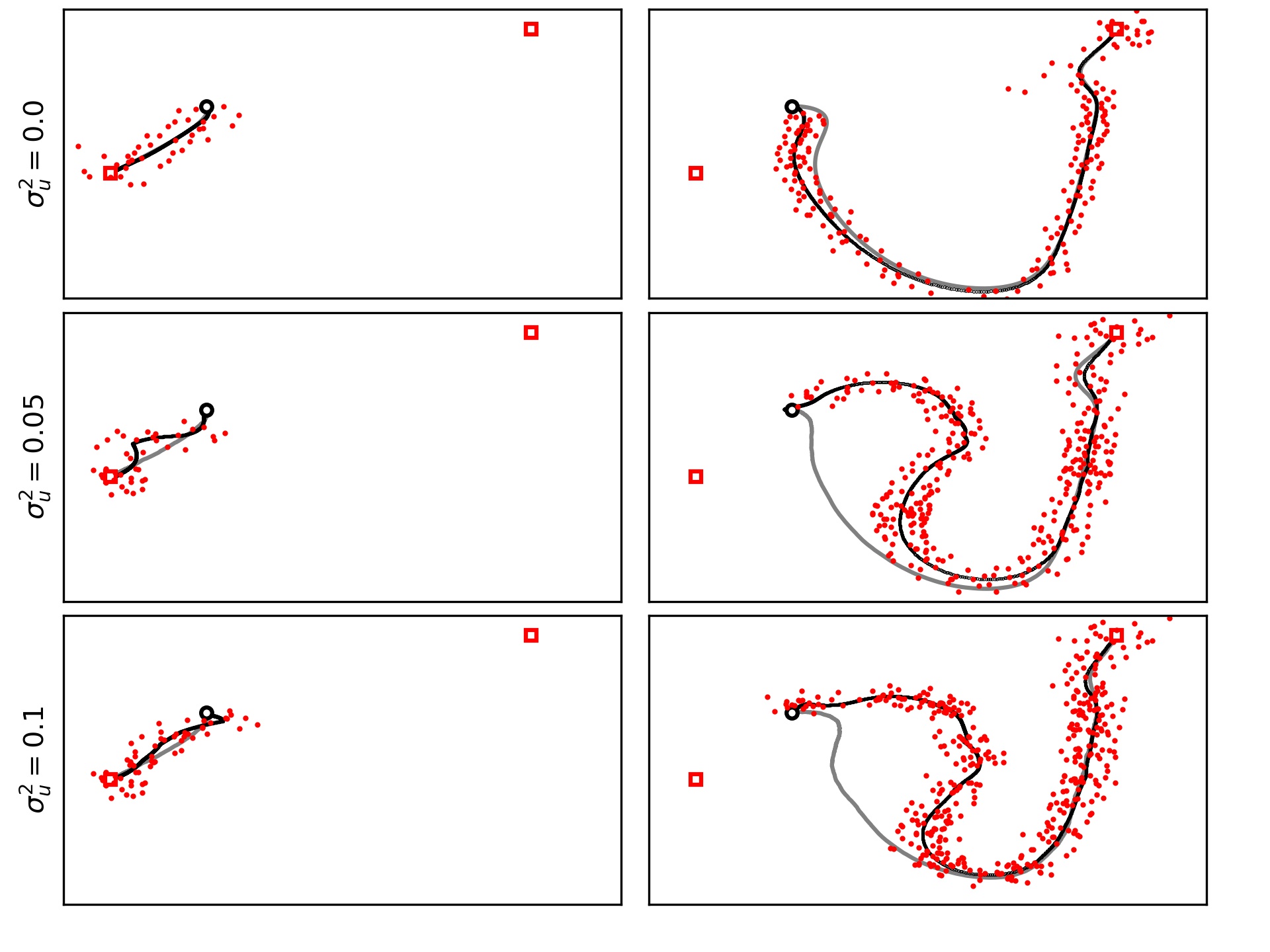}
	\caption{ Comparison between the classical GAD and the aGPR-GAD methods with three different noise levels. The first column and second column show the trajectories (black lines) and the data locations  (red stars) computed by aGPR-GAD method, starting from $m_1$ and $m_2$, respectively. The gray line represents the trajectory computed by classical GAD method.
	The local minimums are marked by white circle with black edge and the saddle points are marked by white square with red edge.} 
	\label{f:eg2_1}
\end{figure}

\begin{table}[ht]
{\footnotesize
  \caption{ Results for non-gradient system example.
  ``Cost'' is based on the number of evaluations of true force field.}  \label{tab:example2}
\begin{center}
  \begin{tabular}{|c|c|c|c|c|c|c|c|c|} \hline
     initial   $\x^{(0)}$ & method & noise level&$\sigma^2_{sur}$  &numerical $\x^{SP}$  &  Cost  \\ \hline
    $(0.59, 0.73)$& GAD& noise-free& $-$ &  $\bm{(1.79,3.30)}$ & $490$  \\
  &  &$\sigma^2_{u} = 0.05$& $-$ &  $(1.80,3.30)$ & $3907$  \\
 &  &$\sigma^2_{u} = 0.10$& $-$ &  $(1.78,3.29)$ & $3650$  \\
 & aGPR-GAD &noise-free& $0.005$ &  $(1.80,3.32)$ & $40$  \\
 &  &$\sigma^2_{u} = 0.05$& $0.007$ &  $(1.79,3.30)$ & $60$  \\
 &  &$\sigma^2_{u} = 0.1$& $0.010$ &  $(1.79,3.30)$ & $40$  \\ \hline
 $(5.87, 6.25)$& GAD& noise-free& $-$ &  $\bm{(1.79,3.30)}$ & $3523$  \\
  &  &$\sigma^2_{u} = 0.05$& $-$ &  $(1.79,3.30)$ & $3624$  \\
 &  &$\sigma^2_{u} = 0.10$& $-$ &  $(1.80,3.30)$ & $4888$  \\
 & aGPR-GAD &noise-free& $0.005$ &  $(1.78,3.37)$ & $210$  \\
 &  &$\sigma^2_{u} = 0.05$& $0.007$ &  $(1.70,3.30)$ & $350$  \\
 &  &$\sigma^2_{u} = 0.10$& $0.010$ &  $(1.84,3.20)$ & $440$  \\ \hline
  \end{tabular}
\end{center}
}
\end{table}

We test the aGPR-GAD method  when the noise size $\sigma^2_u$ varies with results shown in 
Figure \ref{f:eg2_1}.
We find that  aGPR-GAD  can successfully find  the true SP  at tested noise sizes from both initial local minima. 
The true GAD trajectories  are also plotted in Figure \ref{f:eg2_1} for comparison. 
Although the trajectories computed by aGPR-GAD method deviate from the original trajectory in noisy data cases, we find that the our method still works well and
it does  converge to SP. This shows the robustness of  the aGPR-GAD method and 
we attribute this excellent robustness to the intrinsic uncertainty of Gaussian random function.
 Table \ref{tab:example2} summarizes the accuracy and the computational cost at various settings. 
 The $\x^{SP}$ computed by classical GAD  is  the true SP (highlighted in bold).
 We can find that the aGPR-GAD method has a much lower  computational cost (the number of calling the true force field $\b(\x)$).
 
\subsection{Alanine Dipeptide model}
\label{set:alanine}
 In this example, we apply our method to alanine dipeptide, a 22-dimensional Molecular Dynamic model whose collective variables are two torsion angles.
Here, we study the isomerization process of the alanine dipeptide in vacuum at $T=300K$. 
The isomerization of alanine dipeptide has been the subject of several theoretical and computational studies \cite{li2019computing, ren2005transition}, therefore it serves as a good benchmark problem for the proposed method.

\begin{figure}
  \centering
	\includegraphics[width=.25\linewidth]{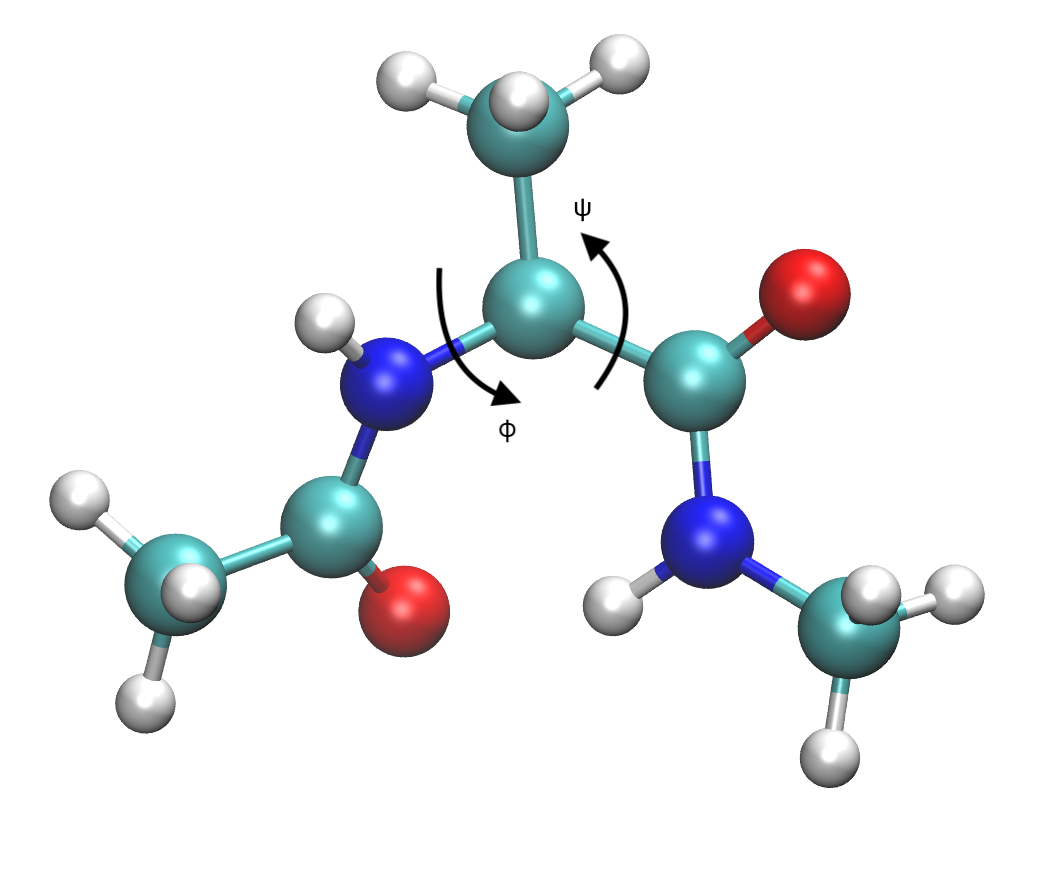}
	\caption{Schematic representation of the alanine dipeptide (CH3-CONH-CHCH3-CONH-CH3).} 
	\label{f:eg3}
\end{figure}

The molecule consists of $22$ atoms and has a simple chemical structure, yet it exhibits some of important features common to bio-molecules. Figure \ref{f:eg3} shows the stick and ball representation of the molecule and two torsion angles $\phi(\x)$ and $\psi(\x)$ which are the collective variables.
The collective variables are  the given functions of $\x$ which are Cartesian coordinates of all atoms.
The free energy associated with $(\phi(\x), \psi(\x))$ is the function depending on $\bm{\nu} = (\nu_1,\nu_2)$ defined as
\begin{equation}
\label{eq:fef}
    F(\bm{\nu}) = -k_BT\ln\Big(A^{-1}\int e^{-\frac{V(\x)}{k_BT} }\times \delta(\nu_1 - \phi(\x))\times\delta(\nu_2 - \psi(\x))d\x \big),
\end{equation}
where $A=\int_{\mathbb{R}^d} e^{-\frac{V(\x)}{k_BT} } d\x$, $T$ is temperature, $k_B$ is a constant, $\delta(\cdot)$ refers the Dirac-delta function, $V(\x)$ refers the potential energy function
of all atom's position  $\x\in \mathbb{R}^d$.
 
The free energy landscape of the alanine dipeptide model is obtained by restrained simulation of sampling  the Gibbs distribution $e^{-V(\x)/k_{B} T}$ with the  fixed two torsion angles ${\bm{\nu}}=(\phi, \psi)$, based on \eqref{eq:fef}.
To obtain the data of free energy $F$ at any given angles, 
we use the package NAMD \cite{NAMD2005} to simulate the Langevin dynamics with the time step size $0.5$ fs.
This program outputs the numerical approximates of $F$ with some noise, which is used in our method to label the data at any location
in the two dimensional angle plane.

\begin{figure}
  \centering
	\includegraphics[width=0.65\linewidth]{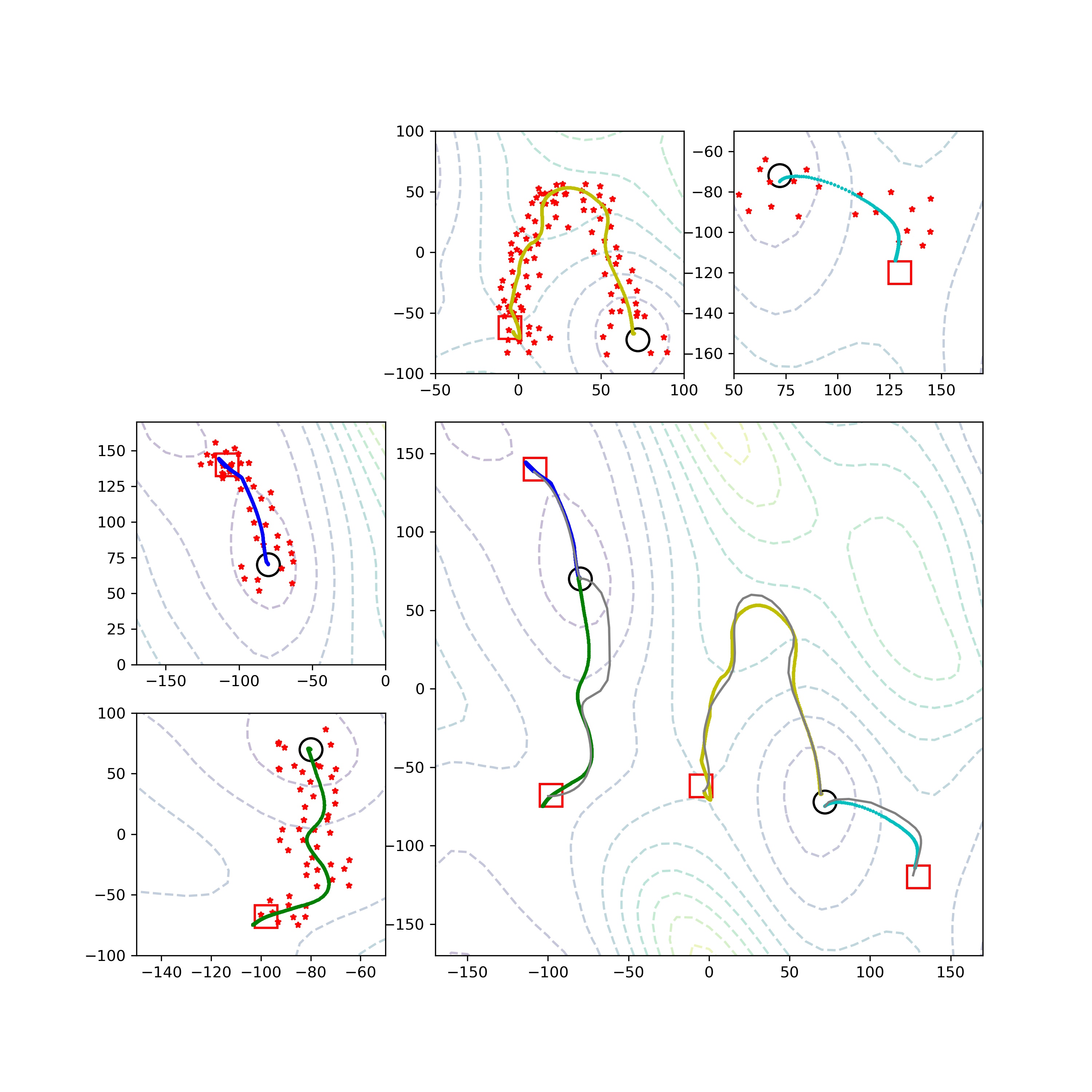}
	\caption{alanine dipeptide example. Two meta-stable conformers and four transition states   are marked as black circles and red squares, respectively. The free energy landscape are shown in contour lines. The gray and colored curves represent the trajectories computed by classical GAD and aGPR-GAD, respectively. The red star points in the surrounding  subfigures 
	highlight the final data locations selected  by our aGPR-GAD method.} 
	\label{f:eg3_sp}
\end{figure}

The molecule has two meta-stable conformers $C_{7eq}$ and $C_{ax}$ located around $(-85^{\circ}, 75^{\circ} )$ and $(72^{\circ}, -75^{\circ})$, respectively. 
The contour plots in Figure \ref{f:eg3_sp} shows the  free energy landscape with respect to torsion angles $\phi$ and $\psi$, precomputed by a very large samples.  
In the figure we marked four TS locations (red squares) and these meta-stable states $C_{7eq}$ (left circle) and $C_{ax}$ (right circle).
Our goal is to find out these transition states, by starting from either $C_{7eq}$ or $C_{ax}$.
In aGPR-GAD method, we use the uncertainty threshold value $\sigma^2_{sur}=1e^{-4}$, $T = 300$, $n=20$, $\Delta t =10$ in both Algorithm\ref{alg:gad-gpr1} and Algorithm\ref{alg:optimal_design}, the number of initial data  $N_0 = 10$ and the number of each batch of optimal design locations $N_D=10$. In classical GAD method, we use $\Delta t = 20$.

The trajectories computed by both classical GAD and aGPR-GAD methods are shown in the central panel of Figure \ref{f:eg3_sp},
with the surrounding small panels to highlight the details of designed data points in   aGPR-GAD search of different saddle points. 
The saving of computational cost is  listed in Table \ref{tab:example3}.
As before, the ``cost''    in aGPR-GAD method   is  the number of queries for accessing the free energy function $F$ by using NAMD.
while the ``cost'' in the GAD means  the number of points in trajectory, i.e, the number of GAD iterations (the actual number of calling $F$ then is  at least a multiplier $6$ of  ``cost''  
\footnote{ Due to the lack of analytic expressions of the gradient, the classic GAD   computes the force field and Hessian matrix of free-energy function by central difference scheme, so 
at least six evaluations of free energy value  are necessary for each point.
}).
 As clearly demonstrated by  Table \ref{tab:example3}, the reduction of the computational cost is striking: only $1\%$ to  $2\%$ 
of the total callings to $F$ for the classic GAD is needed in the new method of aGPR-GAD.

\begin{table}[htbp]
{\footnotesize
  \caption{ Results for alanine dipeptide example.}  
 \label{tab:example3}
\begin{center}
  \begin{tabular}{|c|c|c|c|c|} \hline
     initial point $\x^{(0)}$& method & numerical $\x^{TS}$  & cost  \\ \hline
     $(-80, 71)$   & GAD& $(-108, 138 )$  & $1152$ \\
        & aGPR-GAD& $(-108, 138)$  & $40$ \\\hline
    $(-80, 70)$   & GAD& $( -99, -68 )$  & $702$ \\
        & aGPR-GAD& $(-100, -68)$  & $80$ \\\hline
     $(72, -75)$   & GAD& $( 126, -118 )$ & $639$ \\
       & aGPR-GAD& $(127, -114)$ & $20$ \\\hline
     $(70, -67)$  & GAD  & $( -3, -65 )$  & $774$ \\
        & aGPR-GAD& $(-3, -65)$  & $100$ \\\hline
  \end{tabular}
\end{center}
}
\end{table}

\section{Conclusion}\label{conclusion}
In conclusion, we have presented an active learning method to construct sequential surrogate derivative functions for
efficient saddle point calculation of  computation-intensive energy functions.
The search method is based on the gentlest ascent dynamics (GAD) applied to the surrogate functions.
To optimally update the surrogate functions for best prediction performance for the future GAD trajectories,
our active learning method  based on the mutual information  criterion adaptively selects  the query  locations where the energy function are to be evaluated.
 An efficient numerical implementation of our active learning method is proposed.
Most hyper-parameters are automatically trained with  little human intervention.
The effectiveness of our active learning is manifested in the numerical experiments  that  the selected design points are well distributed along the  GAD's future trajectories.
Our three examples demonstrated the competitive performance of the new method.

 The active learning for constructing sequential surrogate model  is now under very rapid development for simulation accelerations  in many scientific  fields.
 Besides the use of Gaussian process regression here,  the Bayesian neural network or ensembles of neural networks
 are good surrogate models with embedded uncertainty structures, and they can handle  a higher dimensional function than the Gaussian process.


\section*{Acknowledgement}
Shuting Gu acknowledges the support of NSFC 11901211 and the Natural Science Foundation of Top Talent of SZTU GDRC202137.  Hongqiao Wang acknowledges the support of NSFC 12101615 and  the Natural Science Foundation of Hunan Province, China, under Grant 2022JJ40567.
   Xiang Zhou acknowledges
  the support of Hong Kong RGC GRF grants 11307319, 11308121, 11318522 and NSFC/RGC Joint Research Scheme (CityU 9054033).
  This work was carried out in part using computing resources at the High Performance Computing Center of Central South University.

\appendix
\section*{Appendices}

\section{kernel matrices in GPR}
\label{set:appgpr}
 
Assume $u(\x)$ is a Gaussian random field with $\dim$-dimension vector input $\x$.
The covariance relationship between $u(\x)$ and $u(\x')$ can be computed by a kernel function $k_{u,u}$.
Benefiting from the Gaussian process property, the first and second order derivatives are also Gaussian processes.
Here we denote $\b=[b_1,\dots,b_d]^\mathrm{T}$ as the gradient vector and $J = [J_{11}, \dots, J_{ij},\dots,J_{dd}]^\mathrm{T}$ as the Hessian matrix of $u$. 
The covariance functions between different random variables are denoted by 
\begin{equation*}
    K_{u,\b}= [K_{u,b_1},\cdots,K_{u,b_d}],
\\ K_{b,b} =
\begin{bmatrix}
K_{b_1,b_1}&\cdots&K_{b_1,b_\dim}\\
\vdots& &\vdots\\
K_{b_\dim,b_1}&\cdots&K_{b_\dim,b_\dim}
\end{bmatrix},
\end{equation*}
\begin{equation*}
    K_{u,J}= [K_{u,J_{11}},\cdots,K_{u,J_{ij}},\cdots,K_{u,J_{dd}}],
\end{equation*}
\begin{equation*}
 K_{\b,J} = 
 \begin{bmatrix}
K_{b_{1},J_{1,1}}& K_{b_{1},J_{1,2}} &  \cdots & K_{b_{1},J_{2,1}} &\cdots &k_{b_{1},J_{\dim,\dim}}\\
\vdots&  & \vdots& & \vdots&\\
K_{b_{\dim},J_{1,1}}& K_{b_{\dim},J_{1,2}} & \cdots& K_{b_{\dim},J_{2,1}}&  \cdots & K_{b_{\dim},J_{\dim,\dim}}
\end{bmatrix},
\end{equation*}

\begin{equation*}
 K_{J,J} = 
 \begin{bmatrix}
K_{J_{1,1},J_{1,1}}& K_{J_{1,1},J_{1,2}} &  \cdots & K_{J_{1,1},J_{2,1}} &\cdots &k_{J_{1,1},J_{\dim,\dim}}\\
K_{J_{1,2},J_{1,1}}& K_{J_{1,2},J_{1,2}} &  \cdots & K_{J_{1,2},J_{2,1}} &\cdots &K_{J_{1,2},J_{\dim,\dim}}\\
\vdots&  & \vdots& & \vdots&\\
K_{J_{\dim,\dim},J_{1,1}}& K_{J_{\dim,\dim},J_{1,2}} & \cdots& K_{J_{\dim,\dim},J_{2,1}}&  \cdots & K_{J_{\dim,\dim},J_{\dim,\dim}}
\end{bmatrix},
\end{equation*} 
where 
\begin{equation*}
\begin{split}
K_{b_i,b_j} &=\frac{\partial^2 }{\partial x_i \partial x'_j} K_{u,u}, \quad
K_{b_j,u} =\frac{\partial }{\partial x_j} K_{u,u}, \quad K_{J_{i,j},b_{e}}=\frac{\partial^2 }{\partial x_i \partial x_j } \frac{\partial }{ \partial x'_e}K_{u,u}\\
K_{J_{i,j},b_{e}} &= K_{b_{e}, J_{i,j}}^\mathrm{T},\quad K_{J_{i,j},J_{e,g}}=\frac{\partial^2 }{\partial x_i \partial x_j } \frac{\partial^2 }{ \partial x'_e \partial x'_g}K_{u,u}, \quad
K_{J_{i,j}, u} =\frac{\partial^2 }{\partial x_i \partial x_j} K_{u,u}
\end{split}.
\end{equation*}
derived from Equation \eqref{eq:app1}.

\section{The SPSA method}
\label{appspsa}

Here we briefly introduce the SPSA method in the context of our specific applications. In each step, the method only uses two random perturbations to estimate the gradient regardless of the problem's dimension, which makes it particularly attractive for high dimensional problems. Specifically, in step $j$, one first draw a $d*N_D$ dimension random vector $\bm{\Delta}_j = [\Delta_{j,1},\dots,\Delta_{j, d*N_D}]$, where $d*N_D = \text{dim}(D)$, from a prescribed distribution that is symmetric and of finite inverse moments. 
The algorithm then updates the solution using the following equations:
\begin{align}
\begin{split}
D_{j+1} &= D_j - a_jb_j(D_j),\\
b_j(D_j) &= \frac{\mathcal{U}^{LB}(D_j + c_j\bm{\Delta}_j) - \mathcal{U}^{LB}(D_j - c_j\bm{\Delta}_j)}{2c_j}\bm{\Delta}_j^{-1},
\end{split}
\end{align}
where $\mathcal{U}^{LB}$ is computed with Equation \eqref{eq:U18} and
\begin{equation}
\bm{\Delta}^{-1}_j = [\Delta_{j,1}^{-1},\dots,\Delta^{-1}_{j,d*N_D}],\quad a_j = \frac{a}{(A+j+1)^\alpha}, \quad c_j = \frac{c}{(j+1)^\gamma},
\end{equation}
where $A, \alpha, c $ and $\gamma$ being algorithm parameters. 
Following the recommendation of \cite{spall1998implementation}, we choose $\Delta_j \sim \text{Bernoulli(0.5)}$ , $A=100, \alpha = 0.602, c=1$ and $\gamma=0.101$ in this work.

\bibliography{AL_saddle_point}
\bibliographystyle{plain} 
\end{document}